\documentclass{article}
\usepackage[T1]{fontenc}
\usepackage[numbers]{natbib}
\usepackage{cite}
\usepackage[ruled,vlined,linesnumbered]{algorithm2e} 
\usepackage[T1]{fontenc}
\usepackage[utf8]{inputenc}
\usepackage{enumitem}
\usepackage{hyperref}
\usepackage{url}
\usepackage{booktabs}
\usepackage{multirow}
\usepackage{colortbl}
\usepackage{graphicx}
\usepackage{xcolor}
\usepackage{amsmath,amssymb,amsthm}
\usepackage{placeins}
\usepackage[most]{tcolorbox}

\newtheorem{proposition}{Proposition}
\newtheorem{definition}{Definition}
\newtheorem{assumption}{Assumption}

\title{Fresh Memory, Stale Plans: Derivation Currency for Distributed LLM-Agent Memory}

\author{%
\begin{tabular}{c}
Evan Chen$^{1}$, Shiqiang Wang$^{2}$, Christopher G. Brinton$^{1}$\\[0.5em]
$^{1}$Purdue University\\
$^{2}$University of Exeter
\end{tabular}%
}

\newcommand{\planfence}{\textsc{PlanFence}}
\newcommand{\runinhead}[1]{\par\noindent\textbf{#1}\enspace\ignorespaces}
\definecolor{planfencefill}{HTML}{DDEDEA}

\begin{document}
\maketitle

\begin{abstract}
A large language model (LLM) agent that inherits a plan through shared memory can hold the latest requirement yet act on a plan derived from an older one: fresh memory, stale plan. Freshness checks miss this failure because they compare local copies with current state (observation currency) rather than the inputs the plan was derived from (derivation currency). \planfence{} makes derivation currency checkable after a handoff. Stored plans carry exact links to their recorded inputs; before a protected action, \planfence{} follows those links to an action-specific dependency frontier, asks each input's owner for its current head, refreshes what changed, and allows one replan before blocking. Application code supplies the links and declares the scope; no shared memory service is required. Holding the native S-Bus validator fixed, supplying inherited input versions raises detected handoff conflicts from 0/30 to 30/30: retained evidence is the missing ingredient. In 30 live five-agent workflows with a revision inserted after planning, a freshness-only executor acts on the stale plan every time, whereas \planfence{}, like a centralized-lineage baseline that requires a shared store, completes all 30 correctly. In matched replay under emulated LTE traces, \planfence{}'s stall stays within 143--237\,ms per action across a 64$\times$ range of update rates while per-update synchronization grows from 52 to 1{,}196\,ms; synchronization is cheaper only at the lowest tested rates. Scoping queries to the declared dependencies holds traffic at 8.1\,KiB per action, a tenth of all-key validation at 128 keys; at full scope the two tie.
\end{abstract}

\section{Introduction}
LLM agents working on long-horizon tasks need to preserve progress across repeated planning and execution steps~\citep{young2025harnesses}, and distributing this work connects an agent near a local tool with agents hosted on other devices or servers. Once agents sit on edge devices and cellular links, however, every round of coordination has a latency cost, so task state must stay usable while agents update it independently.

Multi-agent frameworks organize collaboration through roles~\citep{autogen2023,metagpt2023,camel2023,chatdev2024}, while agent-memory systems preserve information across interactions~\citep{multiagentmemory2026,collaborativememory2025,mirix2025}. Shared memory therefore holds decisions as well as facts, and a stored plan can outlive its creator's interaction and be inherited by another agent~\citep{pal2026swarmworld}, a \emph{handoff}. That executor, however, inherits the plan without the observations behind it.

\begin{figure}[t]
\centering
\includegraphics[width=0.98\textwidth]{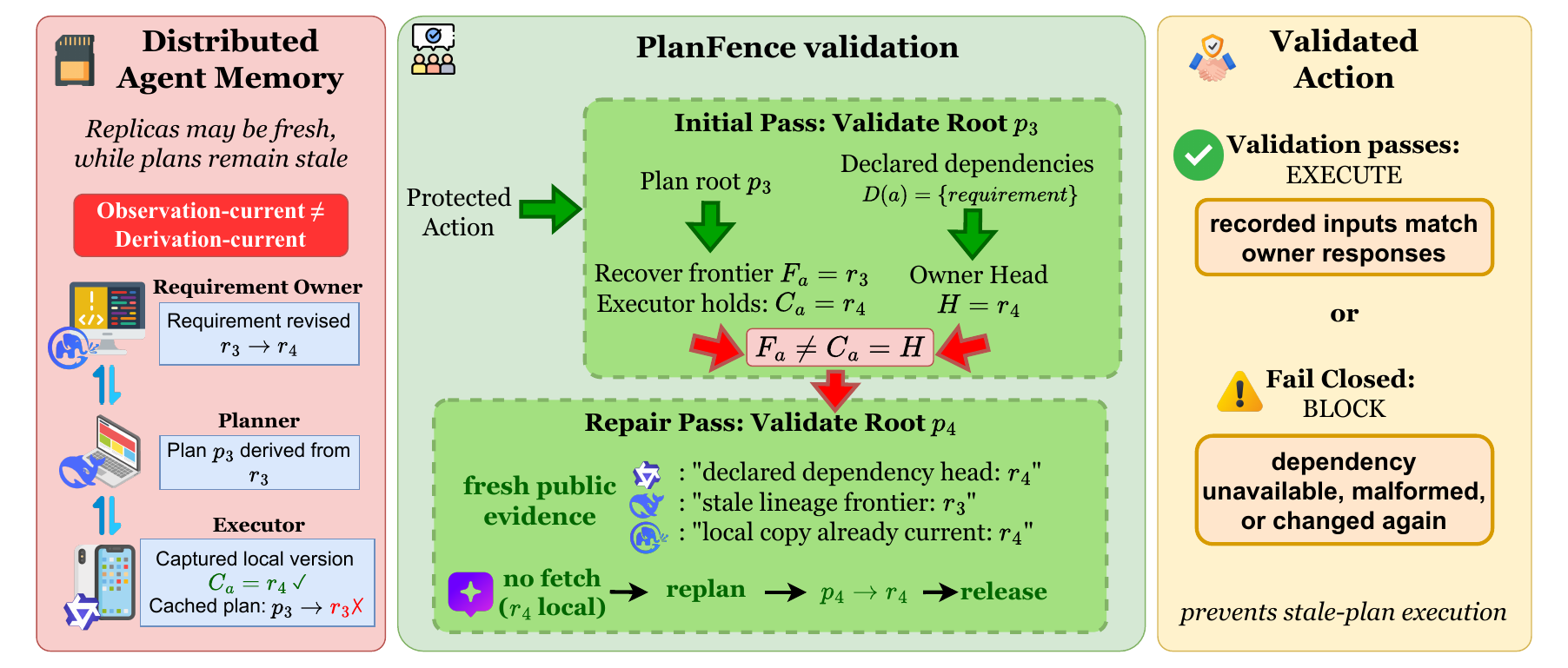}
\caption{Fresh memory, stale plan. A planner derives $p_3$ from requirement $r_3$ and stores it with an exact parent link; the owner then revises the requirement to $r_4$. A later executor retrieves the latest plan ($p_3$) and requirement ($r_4$), so its memory is current, yet the plan's recorded input is $r_3$ (left). \planfence{} compares three views of each declared key: what the plan used ($F_a=r_3$), what the executor holds ($C_a=r_4$), and what the owner has now ($H=r_4$); the mismatch triggers one replan on $r_4$, revalidated before release (center, right). Missing or conflicting evidence blocks instead; content is fetched only when the executor's copy is behind the owner's head.}
\label{fig:stale-plan}
\end{figure}

Consider a planner that stores a shipping plan $p_3$ based on requirement $r_3$, after which a cancellation creates $r_4$ without replacing $p_3$. Another agent then retrieves the latest stored plan $p_3$ and the latest requirement $r_4$, yet still ships the order by following $p_3$: both retrieved objects are the latest versions of their keys, but the plan rests on an obsolete requirement (Figure~\ref{fig:stale-plan}). We call issuing an action in this state \emph{stale-plan execution}, because refreshing memory has not refreshed the decision stored in it.

Checking recorded inputs before an action or commit is established practice~\citep{kung1981occ,khan2026sbus,santosgrueiro2026temporary}, but for an inherited plan the relevant versions may have been read by another agent in an earlier interaction. Existing interfaces can validate supplied versions, so the later executor must first obtain the versions behind this particular plan, and obtaining current versions to compare against requires coordination. Synchronizing after each update communicates between actions, whereas checking when an action is ready puts communication on its execution path, so the resulting cost depends on how often inputs change and how much shared state the action uses. We ask:

\begin{quote}
\emph{How can a later executor validate the inputs behind an inherited plan, and when does checking those inputs at action time reduce coordination?}
\end{quote}

We distinguish the executor's current local copies from the input versions recorded in the plan's derivation. Exact parent links preserve those recorded inputs across handoffs, and an action-specific \emph{dependency frontier} identifies the versions to check, so \planfence{} can reconstruct this frontier and query the corresponding owners when a protected action is ready. Application code supplies the parent links and declares the dependency scope; a mismatch then requests a replacement plan, while missing evidence or an unresolved mismatch blocks execution. We then examine the costs of obtaining current versions at different times and over different scopes. Our contributions are:
\begin{itemize}[leftmargin=*,itemsep=1pt,topsep=2pt]
\item \textbf{Derivation currency and the dependency frontier (Section~\ref{sec:problem}).} We separate whether the executor's memory is current from whether the plan's inputs were current, and define an action-specific frontier that recovers a plan's recorded input versions from the parent links it already carries. The frontier is invariant to later local updates and owner advances, so refreshing memory can never mask a stale plan, and agreement with owner heads is sound for every covered input (Proposition~\ref{prop:soundness}).
\item \textbf{\planfence{} and the missing-evidence result (Sections~\ref{sec:method} and~\ref{sec:rq1}).} A fail-closed validator that reconstructs the frontier when a protected action is ready, checks it against independently owned stores, refreshes changed inputs, and permits one replan before blocking, with no shared memory service and without asking the model what it depended on. With a native validator held fixed, supplying inherited input versions turns 0/30 detected handoff conflicts into 30/30; in live five-agent workflows \planfence{} repairs every inserted revision that a freshness-only executor acts on.
\item \textbf{When action-time checking pays (Sections~\ref{sec:cost} and~\ref{sec:setup}).} A first-order model and matched replay under LTE traces locate two regimes: synchronization is cheaper while changes are rarer than actions, whereas \planfence{}'s stall is nearly flat in the update rate and 1.5--7.1$\times$ below the better of synchronization and a central service once updates outpace actions; scoping queries to declared dependencies keeps traffic flat as unrelated state grows, shrinking to a tie at full scope. A learned selector does not beat this one-threshold rule under shift (Table~\ref{tab:safe-router}).
\end{itemize}

\section{Related Work}
\label{sec:related}

\textbf{Provenance and input validation.}
Data provenance records the inputs behind derived objects~\citep{buneman2001provenance}, and optimistic concurrency control (OCC) checks recorded reads before commit~\citep{kung1981occ}. In the agent setting, S-Bus reconstructs session observations and also accepts explicitly declared read sets across subsystem boundaries~\citep{khan2026sbus}, while MemTX records derived-from relations and propagates invalidation through derived records and effects~\citep{li2026memtx}. Our contribution lies in where the read set lives: OCC and S-Bus bind it to the transaction or session that performed the reads, and MemTX binds derived-from edges to a service that walks descendants forward, whereas \planfence{} binds it to the artifact as exact parent links, so any later holder of the plan can reconstruct the action's inputs without the creator's session, a shared registry, or a descendant index. Table~\ref{tab:rq1}(a) isolates this difference: same S-Bus validator, different evidence.

\textbf{Plan monitoring and action-boundary safeguards.}
Rationale-based monitoring selects environmental features relevant to planning decisions~\citep{veloso1998rationale}, and execution monitoring checks continued plan validity and optimality~\citep{fritz2007monitoring}. For LLM agents, SyncPlan combines explicit synchronization with adaptive replanning in dynamic multi-agent environments~\citep{you2026syncplan}, while StateAuditor addresses responses that still rely on an outdated premise even when updated state is available~\citep{sun2026stateauditor}. Agentic TOCTOU concerns changes between checking and using external state~\citep{lilienthal2025mindgap}; in the same spirit, CommitGuard tracks authority through derived state to a durable effect~\citep{santosgrueiro2026temporary}. \planfence{} sits upstream of these: it establishes, deterministically and without a shared service, that an inherited plan's recorded inputs are still the owners' current versions, and thus neither judges semantic dependence (StateAuditor) nor couples the check to the effect (CommitGuard), while replanning only when a declared input changed rather than on a schedule (SyncPlan).

\textbf{Distributed state and agent memory.}
Registers, replication, and consistency policies offer different guarantees and costs for exposing shared state~\citep{attiya1995abd,demers1987epidemic,shapiro2011crdt,kraska2009rationing,li2012redblue,terry2013pileus,bailis2012pbs}, while agent frameworks structure interaction~\citep{zhuge2024gptswarm,zhang2025maas} and memory systems support retrieval and sharing~\citep{memgpt2023,amem2025,mem0_2025,collaborativememory2025,mirix2025}. SwarmWorld further shows that persistent artifacts are inherited and executed~\citep{pal2026swarmworld}; since retaining an artifact and validating its inputs are distinct responsibilities, \planfence{} supplies the validation step that inheritance needs. Appendix~\ref{app:related-extended} compares the closest systems.

\section{Derivation Currency in Shared Agent Memory}
\label{sec:problem}

Consider agents $\mathcal A=\{a_1,\ldots,a_N\}$ that keep local copies of shared public state, where each requirement, decision, or plan has a semantic key $x\in\mathcal X$ and immutable versions with distinct record IDs. The application assigns each key a single writer, its authoritative owner $o(x)$, whose store holds the key's current version, its \emph{head}, while other agents hold replicas. Derived records in turn retain exact parent IDs identifying their recorded public inputs, whereas prompts and hidden reasoning remain local.

For a \emph{protected action} $a$, one whose external effect the application gates behind validation, let $L(a)$ be the plan roots supplying its arguments and $D(a)\subseteq\mathcal X$ the dependency keys declared by the application. When a validation pass begins, the protected-action wrapper (the application code that gates the action) captures the latest local ID $C_a(x)$ for each declared key, while $F_a(x)$ denotes the recorded input version recovered from the plan, as defined below, and $H(x)$ the head returned by owner $o(x)$ at that owner's read. Owners may be read at different times, but $C_a$ stays fixed during the pass.

\begin{definition}[Observation and derivation currency]
\label{def:currency}
A pending action is \emph{observation-current} if $C_a(x)=H(x)$ for every $x\in D(a)$. It is \emph{derivation-current} if its frontier is defined and
\begin{equation}
F_a(x)=H(x)\qquad\forall x\in D(a).
\label{eq:current-input}
\end{equation}
With a defined frontier, an observation-current candidate that fails \eqref{eq:current-input} is in a stale-plan state. Issuing an action in that state constitutes stale-plan execution. An undefined frontier instead means that validation evidence is unavailable or conflicting.
\end{definition}

In the shipping example, $C_a(x_{\rm req})=\mathrm{id}_4=H(x_{\rm req})$ but $F_a(x_{\rm req})=\mathrm{id}_3$, because installing $r_4$ leaves the recorded parents of $p_3$ unchanged. Informally, then, each declared key has three views: what the plan used ($F_a$), what the executor currently holds ($C_a$), and what the owner has now ($H$). Observation currency is $C_a=H$, derivation currency is $F_a=H$, and the implementation releases only when all three agree.

\textbf{Neither notion implies the other.} The shipping example is observation-current but not derivation-current; an executor holding a plan derived from $r_4$ whose local copy is still $r_3$ is the converse (Appendix~\ref{app:proofs} gives both constructions). The two cases separate stale planning inputs from stale captured content, so our implementation requires both views to match the owner responses and conservatively requests a replacement when either differs (Section~\ref{sec:method}).

\textbf{The dependency frontier.}
An input need not be an immediate parent: $p\rightarrow v\rightarrow r_3$ records a plan based on a review of a requirement, so if the requirement key is declared, traversal reaches $r_3$ and stops that branch. Conversely, a plan based on $r_4$ stops at $r_4$, even if $r_4$ links to $r_3$ as revision history.

For each $x\in D(a)$, let $B(x)$ contain the distinct IDs of records of key $x$ reachable from $L(a)$ along parent paths whose vertices before the endpoint all have keys outside $D(a)$; a declared-key root is itself a boundary, and traversal stops at the first declared key on each branch while continuing the others. The frontier is then $F_a(x)=i$ when $B(x)=\{i\}$ for every declared key, whereas empty or conflicting boundary sets leave it undefined.

\begin{assumption}[Recorded-input contract]
\label{as:all}
Record IDs bind immutable content, keys, and parent lists. Parent links faithfully identify the public inputs recorded for each derivation. The declared scope covers every action-relevant public input boundary, and each such boundary version is reachable from $L(a)$ along a parent path with no declared key before its endpoint. Required ancestry is finite, available, and retained during reconstruction. Owners are benign; responses are authenticated and record identity and integrity are checked.
\end{assumption}

Coverage concerns input boundaries rather than the revision history behind a reached boundary, so a required input reachable only behind another declared key is not covered. In the controlled adapter, application code records the requirement version supplied to the role call and declares that requirement key; the contract thus asks the application to log, not to infer, since the parent links are simply the record IDs the planner placed in the model's context. Declaring every supplied record is safe with respect to \eqref{eq:current-input}, because it can only add replans or blocks, whereas an undeclared input leaves its changes unchecked; whether the model semantically \emph{used} a record is a question for semantic auditing~\citep{sun2026stateauditor}, complementary to the deterministic check here.

\begin{proposition}[Frontier invariance and conditional soundness]
\label{prop:soundness}
For fixed $L(a)$ and $D(a)$, once the required ancestry is available and $F_a$ is defined, installing additional versions or advancing owner heads does not change the recorded frontier. Under Assumption~\ref{as:all}, if \eqref{eq:current-input} holds, every covered recorded input matches its owner's returned version in that pass. A covered input differing from that response prevents release of that candidate.
\end{proposition}

In particular, refreshing local memory can never mask a stale plan. Validation nevertheless establishes neither a common multi-owner snapshot nor currentness at the external effect; what it does is shrink the unprotected window to the interval between the last owner response and the effect, which Appendix~\ref{app:replan-overlap} measures. Appendix~\ref{app:proofs} gives the proofs and Appendix~\ref{app:validation-assumptions} specifies failure handling.

\section{Validating Derivation Currency with \planfence{}}
\label{sec:method}

\planfence{} checks an inherited plan against \eqref{eq:current-input} by recovering the versions behind the plan separately from the executor's latest observations. Owner queries then supply the current versions for comparison, so the executor does not need the creator's interaction history.

\textbf{Constructing the frontier.}
Application code attaches the exact IDs of the public inputs whenever it stores a derived record, so a later executor can start from the roots $L(a)$ and follow the parent links until it reaches the keys in $D(a)$. Algorithm~\ref{alg:planfence}, given in Appendix~\ref{app:validation-assumptions}, follows every branch up to a declared key, using a worklist and a visited set so that a shared record is expanded once, and it continues after finding a version for every key because another branch may contain a conflicting version. Missing records encountered before or at a boundary block validation, whereas older records behind a reached boundary need not be present. With a single declared key and a direct parent, the setting of our cost measurements, the frontier is simply the plan's recorded parent and the check is one ID comparison; the set-valued boundaries and multi-branch traversal matter for wider scopes and multi-hop derivations such as $p\rightarrow v\rightarrow r_3$, and the guarantee in Proposition~\ref{prop:soundness} does not depend on depth.

\textbf{Validation and bounded recovery.}
Because the executor may already hold a changed input, the wrapper captures the local record ID $C_a(x)$ for each declared key, and \planfence{} then queries the owners concurrently for $H(x)$ and releases the action if the recovered inputs, the captured local versions, and the owner responses agree, i.e., $F_a(x)=C_a(x)=H(x)$ for every $x\in D(a)$. If $H(x)\ne C_a(x)$, the validator fetches and verifies the exact owner-reported record and installs it locally, whereas if only the plan is outdated, so that $F_a\ne C_a=H$, the current content is already present and no fetch is needed. Either mismatch returns a replan request together with the owner-reported IDs, whereupon the calling workflow produces a replacement plan from the refreshed inputs, records its parents, and submits its new roots for another check. Only a single replan is allowed, so a further mismatch or unavailable evidence blocks the action.

\section{Coordination Costs of Derivation Validation}
\label{sec:cost}

The same recorded-input check can obtain head information after updates, through a shared service, or when an action is ready. We give a first-order account of announcement and owner-head exchange work in framed bytes to name the regimes these choices create. It models communication components, not service cost or end-to-end latency, and yields two qualitative predictions that Section~\ref{sec:setup} tests: a crossover in the update rate and linear growth of unrelated entries with the keyspace.

Let $\rho$ be owner-head changes per protected action, $N$ the number of agents, $M$ the distinct owners represented in the full keyspace, and $d=|D(a)|$. Measure additive costs in framed metadata bytes: $c_h$ is one announcement to one recipient, $c_r$ the fixed framing and request--response overhead of one query, and $c_e$ its per-head entry cost, without double counting. Let $\nu$ be a fixed modeled number of complete owner-query passes per protected action, matched between the scoped and all-key policies. With fixed dependency scope,
\begin{equation}
W_{\rm sync}\approx\rho(N-1)c_h,\qquad
W_{\rm scoped}\approx\nu d(c_r+c_e),\qquad
W_{\rm all}\approx\nu\bigl(Mc_r+|\mathcal X|c_e\bigr).
\label{eq:work}
\end{equation}
The announcement convention is one delivery to each of the other $N-1$ agents, scoped validation uses per-key queries, and all-key validation batches one query per owner. The model excludes inference, local processing and ancestry traversal, content fetching, local-store operations, and failed-operation behavior, while a shared service additionally incurs update commits and action-time reads.

\textbf{Timing.}
For fixed parameters with $N>1$ and positive $c_h$, the modeled query work is below announcement work when
\begin{equation}
\rho>\rho^*=\frac{\nu d(c_r+c_e)}{(N-1)c_h}.
\label{eq:crossover}
\end{equation}

\textbf{Scope.}
For matched pass multiplicity, $W_{\rm all}-W_{\rm scoped}\approx\nu[(M-d)c_r+(|\mathcal X|-d)c_e]$, derived as \eqref{eq:scope} in Appendix~\ref{app:validation-assumptions}: the unrelated-entry term grows linearly with key count, whereas at full scope, $d=|\mathcal X|$, only the request term $\nu(M-|\mathcal X|)c_r\leq0$ remains, so batching can win there.

Measured synchronization traffic grows with $\rho$ and $N-1$ while scoped traffic stays at 8.1\,KiB, as Tables~\ref{tab:cadence-appendix} and~\ref{tab:team-appendix} show; stall is non-additive in bytes because queries run concurrently, so the byte model orders traffic while sequential exchanges order stall, and neither is fitted.

\section{Experiments}
\label{sec:setup}

Section~\ref{sec:rq1} asks whether a later executor can validate the inputs behind an inherited plan, first with a native validator held fixed while only its evidence varies, then with live repair. Sections~\ref{sec:rq2} and~\ref{sec:rq3} ask when action-time checking reduces coordination, along the two axes of Section~\ref{sec:cost}: update rate $\rho$ and scope ($d$, $|\mathcal X|$). Replay removes the model call so that stall is attributable to coordination alone.

\subsection{Experimental Design}
\textbf{Live and controlled workflows.}
Five Qwen3.5-35B-A3B agents run in separate role processes, with planner, executor, and auditor roles, across three workflow families: reservation, fulfillment, and deployment. The matched live study compares owner-head freshness, centralized lineage, and \planfence{} on 10 seeds per family, each with a single requirement revision inserted after planning and before the protected action; replicated policies use agent-local public stores, whereas centralized lineage uses a single shared store. Controlled replay instead fixes decisions in 30 public workflow templates while retaining the memory, transport, and validation paths across teams of 3 to 8 agents (one template per family in the team-size and horizon sensitivity settings), so that replanning uses recorded fresh decisions rather than new model calls and policy effects are separated from generation variability. Appendix~\ref{app:implementation} gives the model and harness settings.

\textbf{Baselines.}
To separate memory freshness from plan validation, \emph{Local replica} acts from installed state, and \emph{Owner-head freshness} retrieves the current dependency without checking whether the pending plan used it. Among policies that retain exact input IDs, \emph{Centralized lineage} validates through a shared memory service placed across the same emulated link as the replicas; \emph{Metadata sync ($K=1$)} announces every changed owner head and compares plan inputs with synchronized heads; \emph{Per-key all-key validation} queries every shared key with one request per key; and \emph{Batched all-key validation} queries all shared-key heads in one request per owner, whereas \planfence{} queries only $D(a)$. All derivation-current policies therefore share the same recorded-input evidence and release predicate and differ only in how heads are obtained; in replay they also share a budget of one replan followed by revalidation, after which a remaining mismatch or missing evidence blocks execution. Three further controls, delayed synchronization with $K>1$, and team-size and horizon sensitivity are reported in Appendix~\ref{app:controlled-replay}, and Appendix~\ref{app:related-extended} discusses three alternative baselines and what each would require.

\textbf{Workloads.}
Each replay episode forms plans, changes public requirements, and attempts protected actions under matched policy schedules, with five agents, 8 semantic keys, 64 work units (scheduled agent steps) per episode, and a single declared dependency per action unless stated otherwise. We vary the nominal update target $\rho\in\{0.25,1,4,16\}$, measured in updates per protected action, under loopback and fixed historical AT\&T, T-Mobile, and Verizon LTE traces replayed as per-exchange delays, an edge setting~\citep{polese2025airan}; a trace offset is the start point within a trace. The scope comparison then varies the shared keyspace from 8 to 128 keys at $\rho=4$, and a second comparison widens $d$ at a fixed 8-key state, both against batched all-key validation so that dependency scoping is not credited for avoiding unbatched requests.

\textbf{Metrics.}
An issued action is \emph{invalid} if its recorded inputs fail \eqref{eq:current-input}, whatever its local copies hold; it is \emph{available} if issued rather than blocked, and \emph{complete} if it satisfies the current public requirement. Blocking is never counted as valid execution. \emph{Coordination stall} is the time an episode spends in coordination, on the action path (validation, fetch, revalidation) and in synchronization barriers or central commits, per protected action, with generation excluded, as Appendix~\ref{app:statistics} details. Distributed traffic counts inter-replica and central-service bytes. Cost comparisons pair identical schedules and report medians over workflow templates; a difference below 10\% of the larger median is a tie, and separate campaigns (runs of one configuration executed at different times) are compared only within a table.

\subsection{Evidence After a Handoff}
\label{sec:rq1}

We first want to confirm that stale-plan execution actually arises when nothing validates the plan: that an executor can hold the current requirement and still act on a plan derived from an older one. Thirty model-generated workflows run without any validation confirm it. Memory was mostly fresh: only 3 of the 30 executors held an obsolete local copy of the requirement. The plans were not: 28 derived from the initial requirement, and in 15 of those runs the executor held the revised requirement yet issued the incorrect action from the stale plan. The plans named the requirement in their payload but carried no memory-record parent, so nothing could have validated them. This split between fresh memory and stale plans is the empirical origin of the observation/derivation distinction. Because the plan-staleness measure was defined after inspecting these runs, which Appendix~\ref{sec:live-appendix} classifies in full, we treat it as motivation and test the phenomenon under a matched inserted revision below.

\begin{table}[t]
\centering
\footnotesize
\begin{minipage}[t]{\textwidth}
\centering
\setlength{\tabcolsep}{2.5pt}
\begin{tabular}{@{}lrrr@{}}
\toprule
Validator and evidence & Detected & \shortstack{Invalid\\/ issued} & \shortstack{Valid completion\\/ scheduled} \\
\midrule
\multicolumn{4}{@{}l}{\textit{Input revised after same-session planning}} \\
S-Bus validator, session log only & 30/30 & 0/30 & 30/30 \\
S-Bus validator + inherited versions (control) & 30/30 & 0/30 & 30/30 \\
\rowcolor{planfencefill} \planfence{} (ours) & 29/30$^{\dagger}$ & 0/29 & 29/30$^{\dagger}$ \\
\midrule
\multicolumn{4}{@{}l}{\textit{Inherited plan, executor retrieves the revised input}} \\
S-Bus validator, session log only & 0/30 & 30/30 & 0/30 \\
S-Bus validator + inherited versions (control) & 30/30 & 0/30 & 30/30 \\
\rowcolor{planfencefill} \planfence{} (ours) & 29/30$^{\dagger}$ & 0/29 & 29/30$^{\dagger}$ \\
\bottomrule
\end{tabular}
\par\smallskip (a) Native S-Bus validator under three evidence configurations, recorded decisions
\end{minipage}
\par\medskip
\begin{minipage}[t]{\textwidth}
\centering
\setlength{\tabcolsep}{2pt}
\begin{tabular}{@{}lrrrr@{}}
\toprule
Method (memory) & \shortstack{Task\\success} & \shortstack{Invalid primary\\/ scheduled} & \shortstack{Successful\\replans} & \shortstack{Redundant\\auditor actions} \\
\midrule
Owner-head freshness (repl.) & 0/30 & 30/30 & -- & 0 \\
Centralized lineage (shared) & 30/30 & 0/30 & 30/30 & 4 \\
\rowcolor{planfencefill} \planfence{} (ours, repl.) & 30/30 & 0/30 & 30/30 & 10 \\
\bottomrule
\end{tabular}
\par\smallskip (b) Live five-agent workflows
\end{minipage}
\caption{Validation after a plan handoff. (a) Evidence ablation with the native S-Bus validator held fixed, 30 trials replaying recorded model decisions, at most one replan; the control row supplies the inherited versions that \planfence{} recovers from parent links. $^{\dagger}$Two trials time out while their replicas start and stay in the denominators (Appendix~\ref{app:validator-comparison}). (b) Live five-agent workflows, one revision inserted after planning, 10 seeds per family; the freshness-only arm refreshes memory only after its executor has generated the action. Redundant auditor actions are extra valid actions after a valid primary.}
\label{tab:rq1}
\end{table}

Table~\ref{tab:rq1}(a) is an evidence ablation that holds the native S-Bus validator~\citep{khan2026sbus} fixed and varies only the evidence it receives. Session tracking detects all 30 changes after same-session planning, but after a handoff the receiving session has observed the revised requirement and not the inherited plan's old input, so it misses all 30 tested inherited conflicts. Supplying the inherited versions through the supported explicit read-set interface restores detection in all 30 cases; that configuration, however, requires the application to carry the creator's read set across every handoff and to validate against one logical registry, whereas \planfence{} recovers the same versions from the plan's own parent links and detects each conflict in its 29 initialized trials per stale scenario. The two are therefore equally safe here and differ in what must cross the handoff and where validation can run. The unchanged-input control in Table~\ref{tab:validator-comparison} separates version validity from task correctness: all three configurations issue no input-version violations, yet only 28/30 recorded decisions complete the task because two are semantically incorrect on current inputs, so derivation currency does not imply semantic correctness.

Table~\ref{tab:rq1}(b) runs the repair path with live model calls. Owner-head freshness issues an obsolete primary action (the task's protected action) in all 30 tasks: its executor generates the action from local state and refreshes the owner head only afterwards, so the revision arrives after the decision is made and a fresh memory cannot repair an already generated action. Centralized lineage and \planfence{} each detect the revision, regenerate the action from the refreshed inputs in one replanning call by the executor role, and complete all 30 without an invalid primary action. A further arm, reported in Appendix~\ref{app:implementation}, refreshes the executor's inputs before it generates: it adapts its primary action in 28 of 30 tasks yet completes none, because the downstream auditor still reads stale state and undoes the correct action, whereas \planfence{} also hands the validated inputs to that role. Under a single revision, then, the two lineage-retaining validators are equally safe, and what \planfence{} changes is where the evidence lives and what validation costs: no shared service sits on the action path, which in replay at $\rho=4$ removes roughly half the stall and traffic, as Section~\ref{sec:rq2} shows.
The single-revision setting has no update during generation, so Appendix~\ref{app:replan-overlap} replays both policies with a post-replan check while owner updates continue during a synthetic replanning delay. At a 2\,s delay and two relevant updates/s both block all 30 trials, and both complete all 30 when the same rate affects only unrelated keys, so churn in the declared inputs, not global memory activity, exhausts the recovery budget. A live study in which the update lands during the replacement model call likewise blocks all nine such attempts per policy, and one issued \planfence{} action sees an input change between validation and the recorded effect, as Proposition~\ref{prop:soundness} allows.

\subsection{Timing of Coordination}
\label{sec:rq2}

\begin{table}[t]
\centering
\footnotesize
\setlength{\tabcolsep}{3.0pt}
\begin{tabular}{llrrrr}
\toprule
Category & Method / variant & \shortstack{Invalid\\/ issued} & \shortstack{Available\\/ scheduled} & \shortstack{Stall\\(ms/action)} & \shortstack{Traffic\\(KiB/action)} \\
\midrule
\multirow{2}{*}{\shortstack[l]{Unsafe\\freshness}} & Local replica & 330/330 & 330/330 & 0.0 & 3.8 \\
 & Owner-head freshness & 330/330 & 330/330 & 151.8 & 7.1 \\
\midrule
\multirow{4}{*}{\shortstack[l]{Derivation-\\current}} & Centralized lineage & 0/330 & 330/330 & 508.6 & \underline{15.6} \\
 & Metadata sync, $K=1$ & 0/330 & 330/330 & 403.4 & 23.5 \\
 & Batched all-key validation & 0/330 & 330/330 & \underline{258.4} & 16.4 \\
\rowcolor{planfencefill} & \textbf{\planfence{} (ours)} & 0/330 & 330/330 & \textbf{230.8} & \textbf{8.1} \\
\bottomrule
\end{tabular}
\caption{Policies at the high-update setting ($\rho=4$, i.e., 42 updates and 11 protected actions per episode) on the AT\&T trace with 8 keys and a new TCP connection per RPC. Bold and underlined costs mark the best and second-best values among policies with no invalid action and full availability; per-key all-key validation and further controls are in Table~\ref{tab:all-controls}. Trace offset 0; Appendix~\ref{app:statistics} gives interquartile ranges and the other campaigns of this configuration.}
\label{tab:causal-headline}
\end{table}

Every action in Table~\ref{tab:causal-headline} is a stale-plan race: at this target a relevant update lands between planning and every protected action, so the freshness-only controls act on obsolete plans 330 times, and the cost columns price the full detect, fetch, replan, and revalidate path for every policy that catches the change. The derivation-current policies each issue 330 current-input actions with full availability, and the non-delayed controls remain valid and available across the aligned 32,700-action grid of Appendix~\ref{app:evidence-completeness}, whereas delayed synchronization admits invalid actions, as Table~\ref{tab:all-controls} shows. Among the safe policies, \planfence{}'s stall is 43\% below metadata sync and 55\% below centralized lineage, while batched all-key validation, which also validates at action time, trails it by 12\% on stall here, by 20--25\% at every tested rate in the contemporaneous four-policy campaign of Appendix~\ref{app:controlled-replay}, and by twice the bytes; the two action-time policies thus separate on scope, which Section~\ref{sec:rq3} measures. For scale, a live role call takes about 0.6\,s at the median (Appendix~\ref{app:implementation}), so at $\rho=16$ metadata sync's 1.2\,s per action is two role calls while \planfence{} stays near 0.23\,s (Table~\ref{tab:cadence-appendix}).

\begin{figure}[t]
\centering
\includegraphics[width=0.88\textwidth]{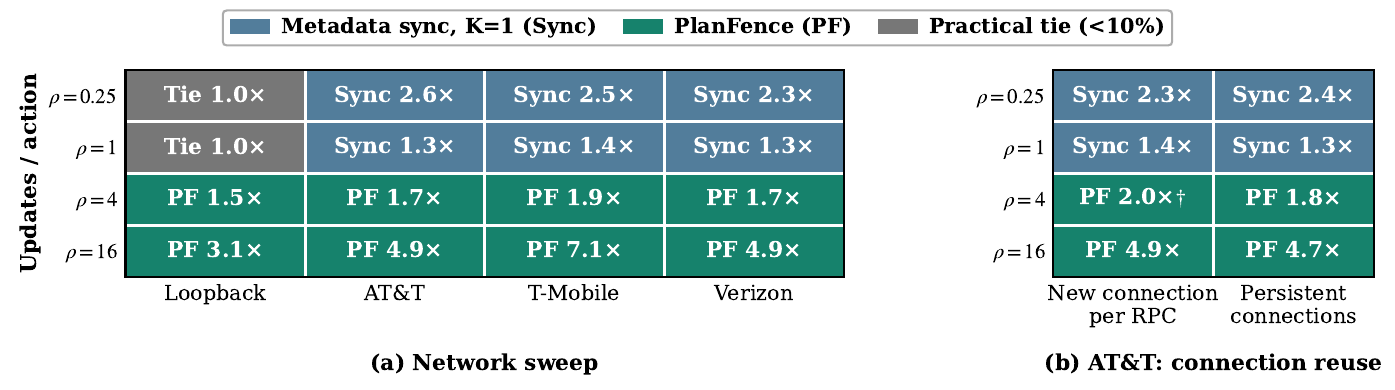}
\caption{Lowest-stall policy per cell among metadata sync ($K=1$), centralized lineage, and \planfence{}; centralized lineage never wins. Cells give the runner-up-to-winner ratio of median stall; gray marks ties (below 10\%). (a) Loopback and three LTE traces; the AT\&T column is tabulated in Table~\ref{tab:cadence-appendix}. (b) AT\&T with new and persistent connections, from the separate campaign of Table~\ref{tab:connection-reuse}; $\dagger$: one of 330 actions blocked. Table~\ref{tab:a2-campaign} adds batched all-key validation at every rate.}
\label{fig:policy-boundary}
\end{figure}

Figure~\ref{fig:policy-boundary}(a) compares metadata sync, centralized lineage, and \planfence{} across loopback and three cellular traces. At $\rho\in\{0.25,1\}$ metadata sync has lower or tied stall; both loopback comparisons are ties, and on the cellular traces at $\rho=0.25$ \planfence{} is the slowest of the three (Table~\ref{tab:connection-reuse}), because it pays an owner round trip even when nothing changed, precisely the regime where announcing rare changes is cheaper. At $\rho\in\{4,16\}$, by contrast, \planfence{} has lower stall in all eight cells, by 1.5$\times$--7.1$\times$ relative to the better of the other two, and the ordering follows from the shape of the curves: across a 64$\times$ range of $\rho$, \planfence{}'s median stall stays within 143--237\,ms per action, one owner round trip per action however often inputs change, whereas metadata sync grows from 52 to 1{,}196\,ms because every change is announced to every agent (Table~\ref{tab:cadence-appendix}), the $\rho$-linear term in \eqref{eq:work}. On loopback, action-time validation is never worse than synchronization, and connection reuse preserves the ordering at all four targets (Figure~\ref{fig:policy-boundary}(b)). The ordering yields the rule metadata sync if $\rho\leq1$ and \planfence{} otherwise, which a learned selector of the coordination policy does not beat under distribution shift (Table~\ref{tab:safe-router}).

\subsection{Scope of Coordination}
\label{sec:rq3}

\begin{figure}[t]
\centering
\includegraphics[width=0.82\textwidth]{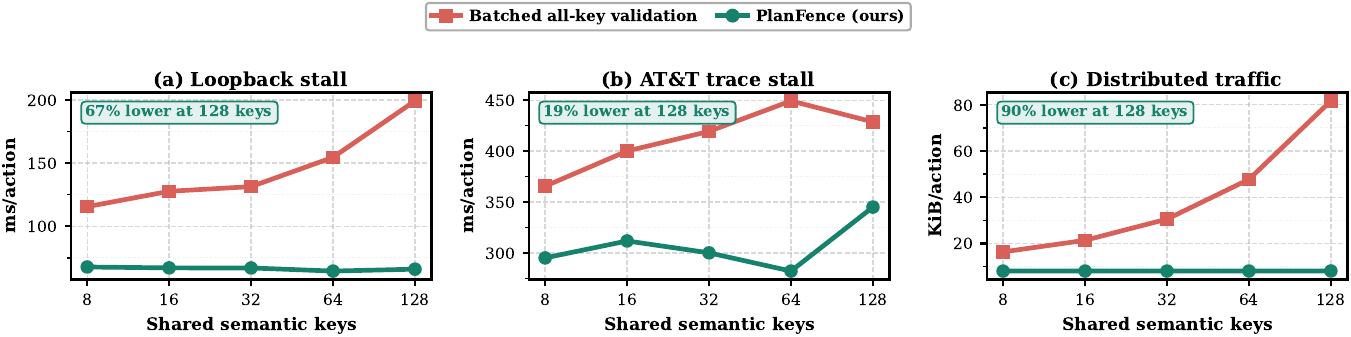}
\caption{Scope against batched all-key validation as the keyspace grows from 8 to 128 keys ($d=1$, $\rho=4$). Loopback points aggregate 30 templates; AT\&T points are medians over three trace offsets (Table~\ref{tab:additional-controls}).}
\label{fig:keyspace-main}
\end{figure}

Figure~\ref{fig:keyspace-main} compares \planfence{} with batched all-key validation as the keyspace grows from 8 to 128 keys, with five agents, one dependency per action, and the same bounded recovery path. Scoped traffic stays near 8.1\,KiB/action while all-key traffic grows from 16.4 to 81.7\,KiB/action, and on loopback stall moves from 115.5 versus 67.8\,ms/action at eight keys to 199.2 versus 66.0\,ms/action at 128 keys (all-key versus \planfence{}). Across three AT\&T offsets, the medians of the per-offset medians are 365.7 versus 295.3 and 428.7 versus 345.2\,ms/action; per-offset reductions range from 5\% (a tie) to 24\% at eight keys and from 19\% to 39\% at 128 keys (Table~\ref{tab:additional-controls}). The byte growth, about 0.54\,KiB per action per key, is exactly the unrelated-entry term in \eqref{eq:scope}.

Table~\ref{tab:fanin} instead fixes the eight-key state and widens the declared scope. At $d=1$ \planfence{} reduces both costs, at $d=2$ stall is a tie while traffic remains lower, and at $d=8$ both costs tie with batching slightly ahead: 400.1\,ms and 34.1\,KiB/action versus 421.7\,ms and 36.8\,KiB/action for \planfence{}. At full scope there are no unrelated entries left to exclude and the scoped policy still issues one request per declared key, so the request-overhead term $\nu(M-|\mathcal X|)c_r\le0$ in \eqref{eq:scope} predicts this direction, and grouping keys by owner would remove it. These measurements therefore bracket the dependency-density boundary, where scoping stops paying: the stall benefit is gone by $d=2$, the traffic benefit between $d=2$ and $d=8$.

\FloatBarrier
\section{Conclusion}
\label{sec:conclusion}

Current local memory does not establish that an inherited plan uses current recorded inputs. The dependency frontier recovers those inputs from the plan's own parent links, and \planfence{} validates them against distributed owners before a protected action, with no shared memory service. The handoff comparison shows that a later executor lacks inherited input evidence, not validator logic; live workflows demonstrate repair after an inserted revision. In matched replay \planfence{}'s cost is nearly independent of update rate and keyspace, whereas synchronization cost grows with churn and fan-out, so synchronization is cheaper only at or below about one update per action. Scoping queries cuts communication while most shared state is unrelated; the benefit disappears, as the model predicts, at full scope.

The contract requires application-supplied dependencies and faithful parent links, and owner reads provide neither a common snapshot nor atomicity with the effect. The live study uses one model family and one inserted revision, and continuing updates are studied mainly under synthetic replanning delays. Future work can examine sustained live generation, retrieval-time parent capture for top-$k$ semantic memories, and changing owner assignments.

\subsection*{AI Use Statement}
In this work, we used a generative AI coding assistant to help implement the experimental software and to help prepare the experiments, under the authors' direction and review; the research questions, the experimental design, and the data analysis are the authors' own. We also used a generative AI assistant to help develop the conceptual framework that distinguishes observation currency from derivation currency, to help formulate the definitions and proposition in Section~\ref{sec:problem} and the coordination cost model in Section~\ref{sec:cost}, and to help interpret the measured results against that model. We have not used generative AI tools to generate synthetic data sets, to clean or reformat data, or to assist in translation, and the remaining tasks with required disclosure are not applicable to this work. Additionally, we used generative AI tools to search for and summarize related literature, to suggest the structure of the paper, and to draft and edit parts of the text for readability. We have reviewed all AI-assisted work, including the AI-assisted code. The formal statements and their proofs were checked line by line by the authors, the cost-model predictions were compared against the measured tables by the authors, and every cited work was verified against its published source. We take responsibility for the final content of this work, including text, claims, and artifacts produced with the aid of generative AI. The language model that acts as an agent inside the evaluated workflows is the object of study and is described in Appendix~\ref{app:implementation}.

\subsection*{Reproducibility Statement}
Section~\ref{sec:problem} and Appendix~\ref{app:proofs} state the formal contract and proofs. Algorithm~\ref{alg:planfence} specifies validation, and Appendix~\ref{app:implementation} describes the model, prompts, storage, transport, workloads, and recovery paths. Appendix~\ref{app:statistics} states the statistical conventions, including how repeated configurations across separately executed campaigns are compared, and Appendices~\ref{app:controlled-replay}--\ref{app:connection-reuse} provide controls and failure accounting. These descriptions distinguish validation-time agreement, availability, and task completion.

\subsection*{Ethics Statement}
The evaluated tasks are controlled reservation, fulfillment, and deployment workflows. The validator checks recorded input versions; it does not establish that an action is authorized or semantically appropriate. Applications with consequential effects require safeguards suited to those effects, including any required coupling between validation and execution.

\bibliography{agentmem}
\bibliographystyle{IEEEtran}

\clearpage
\appendix
\numberwithin{equation}{section}
\numberwithin{table}{section}
\numberwithin{figure}{section}

\section{Extended Related Work}
\label{app:related-extended}

\textbf{Obtaining and retaining validation evidence.}
S-Bus reconstructs an agent's HTTP-observable read set using a session DeliveryLog. Its explicit read-set interface also supports declarations across subsystem boundaries~\citep{khan2026sbus}. Thus a later executor can validate inherited versions if application code supplies them, as the native comparison demonstrates. A session read set can preserve earlier observations even after the local store advances; it must not be equated with the captured latest local IDs $C_a$. \planfence{} reconstructs $F_a$ from persistent ancestry independently of that receiving session. S-Bus validates against a logical registry and supports replicated deployment, whereas \planfence{} queries the owners assigned to the declared public keys. This comparison concerns the evidence and validation architecture, not a measured latency advantage over S-Bus.

\textbf{Derived records and downstream effects.}
MemTX records derived-from edges across beliefs, summaries, shared copies, and tool actions; its commit pipeline checks dependencies, and revocation walks descendants for typed repair~\citep{li2026memtx}. CommitGuard follows authority witnesses through derived plans and other state to the durable-effect boundary, requiring freshness, causal priority, effect binding, and eligibility~\citep{santosgrueiro2026temporary}. Both systems use provenance to govern downstream actions. \planfence{} studies a deliberately minimal version-agreement contract with an action-specific stopping boundary and independently queried owners: no shared bus registry (S-Bus), no descendant index or invalidation fan-out (MemTX), and no effect-side binding (CommitGuard). The price is a guarantee that is conditional on the declared and recorded inputs and ends at the owner responses; the return, measured in Section~\ref{sec:setup}, is a stall reduction of 1.5--7.1$\times$ against synchronization and a central service at high update rates and a tenfold traffic reduction at 128 keys.

\textbf{Behavior after a state update.}
StateAuditor explicitly studies cases where updated state is available but a drafted response still depends on an obsolete premise~\citep{sun2026stateauditor}. Its audit uses model judgments together with deterministic provenance and chronology checks to guide behavioral repair. Our controlled adapter records the versions supplied to planning, and the validator compares record identities without inferring semantic dependence. Classical rationale-based monitoring similarly focuses attention on state relevant to planning~\citep{veloso1998rationale}; the declared frontier supplies a concrete recorded-version boundary for the distributed memory setting studied here.

\textbf{Baselines we did not run.}
Always replanning at action time is derivation-current by construction but pays a regeneration call on every protected action (about 0.6\,s at the median in the matched live study, Appendix~\ref{app:implementation}) and still needs the current inputs, that is, the same owner read; \planfence{} and centralized lineage regenerate only when a declared input changed, exactly once per task in Table~\ref{tab:rq1}(b), and in replay a recorded-decision replan would be free, so the coordination grid cannot price always-replan fairly. A timestamp or version comparison against a synchronized head is metadata sync with $K=1$, and the $K>1$ rows of Table~\ref{tab:cadence-appendix} show what a comparison against a non-current head admits. A model-based judge of whether a plan still fits the requirement~\citep{sun2026stateauditor} needs the superseded requirement to compare against; after a handoff that record is available only if something retained it, so a judge faces the same evidence question and is complementary to the deterministic check.

\section{Proofs}
\label{app:proofs}

\subsection{Constructions for the non-implication remark in Section~\ref{sec:problem}}

Use one declared key $x$ with immutable versions $r_3,r_4$ and distinct IDs $\mathrm{id}_3,\mathrm{id}_4$. Each plan records the ID of the input supplied to its derivation.

For the first direction, derive $p_3$ from $r_3$, then advance the owner to $r_4$ and install $r_4$ at the executor. Retain $p_3$ and its ancestry. A pass captures $C_a(x)=\mathrm{id}_4$, reconstructs $F_a(x)=\mathrm{id}_3$, and receives $H(x)=\mathrm{id}_4$. Hence observation currency holds and derivation currency fails.

For the converse, let the owner remain at $r_4$ and let a planner produce $p_4$ from $r_4$. An executor holding $p_4$ captures $C_a(x)=\mathrm{id}_3$ at pass entry. Before traversal, it receives the previously created $r_4$, making the required ancestry available. Installation does not alter the captured $C_a$. Traversal therefore reconstructs $F_a(x)=\mathrm{id}_4$ and the owner returns $H(x)=\mathrm{id}_4$, while $C_a(x)\ne H(x)$. This construction explicitly separates available ancestry during reconstruction from the earlier local capture; if required ancestry remains unavailable, validation blocks instead.

Neither construction changes an immutable record. \qed

\subsection{Proof of Proposition~\ref{prop:soundness}}

\emph{Invariance.} Fix $L(a)$ and $D(a)$ and suppose all required ancestry is available and retained, with a defined frontier. The boundary traversal visits records according to their immutable keys and parent lists. Its visited set ensures each record is expanded at most once; the accessible ancestry is finite by assumption. Installing additional records does not change the already available records, their IDs, or their outgoing parent links. Advancing owner heads changes only the comparison mapping $H$. Thus the visited boundary sets $B(x)$ and their singleton values $F_a(x)$ remain unchanged. Installing an ancestor that was previously missing can change a local reconstruction from failure to success; it does not alter the plan's intrinsic historical mapping and is outside the premise of an already available frontier. Digest-bound record IDs preclude cycles, and the visited set guards against malformed stores.

\emph{Conditional soundness.} Let $i$ be any covered recorded input boundary of declared key $x$. Coverage supplies a path from a root to $i$ whose vertices before the endpoint all lie outside $D(a)$. Faithful parent links and available ancestry allow the traversal to follow that path without stopping earlier, so $i\in B(x)$. Since the frontier is defined, $B(x)=\{F_a(x)\}$. Under \eqref{eq:current-input}, $i=F_a(x)=H(x)$, which is the ID returned by the corresponding owner in this pass. The argument applies separately to each owner read and does not establish a simultaneous snapshot.

\emph{Non-release of a mismatching candidate.} If a covered boundary ID $i\ne H(x)$, it belongs to $B(x)$ by coverage. Multiple IDs make the frontier undefined and block validation. Otherwise $F_a(x)=i\ne H(x)$, so the implemented release predicate $F_a=C_a=H$ fails. The candidate is blocked or returned for replanning; a new candidate may later pass. Missing ancestry, unavailable owners, or failed integrity checks also prevent release. These conclusions concern covered recorded inputs, not semantic correctness, authorization, or guaranteed completion. \qed

\section{Implementation Details}
\label{app:implementation}

Each experiment run uses separate agent processes communicating with durable, versioned memory over authenticated TCP, and Table~\ref{tab:implementation} gives the complete configuration. Replicated methods use a single SQLite memory process per agent, while centralized lineage uses a single shared SQLite memory process. In replay that service is a central metadata directory placed across the same emulated link as the replicas, so the comparison varies only where heads are validated; in the live study and Appendix~\ref{app:replan-overlap} it stores the public content itself. Public records are immutable JSON objects in durable SQLite stores. Paired policies receive the same initial records and schedules, and network impairments begin only after setup. Infrastructure-only recovery attempts are linked separately in the follow-up studies and do not replace the primary failures in their scheduled denominators. Policy-induced blocks, failures, and timeouts remain in the outcome accounting described for each study. Every retained initialization failure and the single aborted episode occur in the \planfence{} arm, the replicated five-process configuration. The two initialization failures in Table~\ref{tab:validator-comparison} and the six in Table~\ref{tab:replan-overlap} are bootstrap-install RPC timeouts against the 10\,s request timeout (one logged bootstrap response arrives after about 19\,s), before the affected trial's validation begins; the aborted episode in Table~\ref{tab:connection-reuse} is an owner-write RPC timeout during an owner update, not a validation refusal; and the two timeout blocks in Appendix~\ref{app:connection-reuse} are a local validate RPC (\planfence{}, new connections) and a local owner-heads RPC (batched all-key validation, reused connections) that exceed the 10\,s timeout. We did not isolate host or storage causes, we keep every such failure in its denominator, and the separately linked reruns show only that the settings themselves initialize.

Each stored public version carries its semantic key, immutable record ID, authoritative owner, monotone owner sequence, exact parent IDs, record type, content size, and public content or content address. An owner accepts a head update only if the author identity of the record matches the declared owner and the sequence increases, and duplicate owner sequences are conflicts. Head announcements send the metadata envelope, while consumers fetch content only after discovering that their local version is obsolete.

The parent links are stored with the derived record and are not reconstructed from a later prompt or memory snapshot. In the controlled live adapter, a stored role decision names the exact requirement record supplied to that role call. The plan record uses the same link, and a replacement action records the refreshed requirement as its parent. The protected-action wrapper declares the workflow instance's requirement key and its owner. Controlled replay supplies the selected dependency keys and their exact parent IDs when it stores a plan. Neither path asks the model to infer the dependency declaration.

\begin{tcolorbox}[
  enhanced,breakable,colback=blue!3,colframe=blue!55!black,
  title=\textbf{Selected fields of a plan record},fonttitle=\bfseries,
  coltitle=white,colbacktitle=blue!55!black,
  boxrule=0.6pt,arc=1.5mm,left=1.5mm,right=1.5mm,top=2mm,bottom=1.5mm]
\small
\texttt{key}: the workflow's plan key.\quad
\texttt{record\_type}: \texttt{controlled\_plan}.\par
\texttt{record\_id}: $\mathrm{id}_p$.\quad
\texttt{writer\_id}: the authoring replica.\quad
\texttt{writer\_seq}: its sequence number.\par
\texttt{parent\_record\_ids}: $[\mathrm{id}_{r_3}]$.\quad
\texttt{payload}: the public decision and requirement record ID.
\smallskip\par
IDs here are illustrative symbols for the stored content digests, and they are neither task labels nor current-head pointers. The complete record also carries clocks and the remaining public content. Its digest binds that content together with the exact parent IDs, so installing $r_4$ does not modify $\mathrm{id}_p$ or its link to $r_3$.
\end{tcolorbox}

\begin{table*}[ht]
\centering\small
\setlength{\tabcolsep}{5pt}
\begin{tabular}{lp{0.75\textwidth}}
\toprule
Component & Setting \\
\midrule
Live model & Qwen3.5-35B-A3B, temperature 0.2, top-$p$ 0.95, 1{,}024 output-token cap, thinking disabled, no response cache \\
Live team & Five agent processes. Replicated methods use five local-memory processes, while centralized lineage uses a single shared memory process. 8 role units, with an additional executor replanning call when protected replanning is required \\
Storage and transport & Durable SQLite, authenticated TCP, 10\,s replay request timeout, 30\,s controlled-live memory-operation timeout, fail closed on malformed, conflicting, or unavailable owner state \\
Replay workload & 30 workflow templates at the primary setting ($N=5$, $T=64$) and one template per family at the other team sizes and horizons, $N\in\{3,5,8\}$, episode length $T\in\{16,64,256\}$ work units, nominal update targets $\rho\in\{0.25,1,4,16\}$ realized at $T=64$ as $\{0.25,1.0,42/11,14\}$, dependency counts $\{1,2,8\}$ \\
State and network & Compact records up to 4\,KiB, verified artifacts of 256 to 320\,KiB, loopback and pinned historical AT\&T, T-Mobile, and Verizon LTE traces \\
Scaling controls & Semantic key counts $\{8,16,32,64,128\}$, trace offsets 0, 5, and 15\,s, compact per-owner change digests \\
Execution & Live inference uses four H100 GPUs, replay runs use isolated CPU processes, and all paired policies receive identical schedules \\
Metrics & Invalid issued actions, available and completed actions, coordination stall, distributed wire bytes, replans, and remedial actions \\
\bottomrule
\end{tabular}
\caption{Implementation and harness configuration.}
\label{tab:implementation}
\end{table*}

\subsection{Metrics and statistical analysis}
\label{app:statistics}

Coordination stall is computed per episode as the sum of action-path coordination time (validation, fetch, and revalidation) and timed synchronization barriers or central metadata commits, divided by the episode's attempted protected actions, and then summarized as the median over templates. It excludes generation and plan-memory time, counts concurrent requests by their elapsed duration rather than by summing participants, retains the cost of blocked actions in completed episodes, and assigns no latency to an aborted episode. Binary endpoints are reported as raw counts, with invalid issued actions separated from blocked and completed actions. Repeated actions share workflow templates and schedules, so we do not interpret the aggregate zero-event counts as independent deployment trials.

Cost comparisons pair identical schedules and report medians over workflow templates, with interquartile ranges where a table lists them. All comparisons are descriptive. The primary and keyspace analyses use 10{,}000 two-stage bootstrap draws that resample the three workflow families with replacement and then paired template differences within each sampled family; the dependency-density analysis uses 2{,}000 such draws, and the connection-reuse follow-up uses 2{,}000 draws that resample paired templates (Appendix~\ref{app:connection-reuse}). With three families the outer stage is coarse, so these intervals describe sensitivity to family composition and within-family template variation, not population-wide uncertainty. We regard a reduction of at least 10\% of the larger median stall or traffic as practically meaningful if validity does not weaken and availability falls by no more than five percentage points; the threshold is a reporting convention, not a calibrated noise floor.

\textbf{Repeated configurations across campaigns.} The $\rho=4$, eight-key AT\&T configuration at trace offset zero appears in the primary campaign (Table~\ref{tab:causal-headline}), the density and offset campaigns (Tables~\ref{tab:fanin} and~\ref{tab:additional-controls}), the connection-reuse follow-up (Table~\ref{tab:connection-reuse}), and the replication batches of Appendix~\ref{app:a2-a1}. Schedules are matched and traffic is byte-identical across all of them, and every campaign except the reuse arm opens a new connection per RPC. \planfence{}'s median stall is 224--257\,ms per action in every campaign. Batched all-key validation's is 258--265\,ms in the three campaigns on the earlier harness revision and 326--342\,ms in the five batches on the current revision, which logs connection lifecycle events and fsyncs each event append; its connection counts also differ slightly between campaigns. We have not isolated instrumentation from host and timing variability, so we do not compare absolute stall across revisions. The ordering \planfence{} below batched all-key validation holds in all eight campaigns, by 10.7\% in the primary campaign and by 15--25\% in the others, and within the primary campaign the interquartile ranges over templates (223.1--240.6 versus 255.5--260.5\,ms) do not overlap.

\subsection{Controlled live prompting and tool grounding}

Each role receives the same short system instruction with its role and workflow family substituted. The user content is a stable JSON object assembled from only the role's local public state and prior public handoffs. The tool schema then enumerates every permitted identifier, action name, and argument object.

\begin{tcolorbox}[
  enhanced,breakable,colback=blue!3,colframe=blue!55!black,
  title=\textbf{Controlled live-agent prompt},fonttitle=\bfseries,
  coltitle=white,colbacktitle=blue!55!black,boxrule=0.6pt,arc=1.5mm,
  left=1.5mm,right=1.5mm,top=2mm,bottom=1.5mm]
\small
\textbf{System.} \textit{You are \texttt{<WORK\_UNIT>} in a controlled \texttt{<FAMILY>} workflow. Use only supplied evidence. Call exactly one supplied tool. Do not invent IDs or hidden state. The tool schema enumerates the complete allowed IDs and actions for this decision.}

\smallskip
\textbf{User object.} Local requirement and its application-level request ID, an opaque digest of the local requirement revision, up to five locally eligible catalog candidates, prior public handoffs, and, for the auditor, the externally visible action outcome.

\smallskip
\textbf{Grounded tool.} Exactly one function call is required. Candidate IDs, requirement revision, action name, and complete action arguments are finite enumerations constructed from the local query. Additional fields are rejected.
\end{tcolorbox}

A single public correction is allowed after a schema violation. The correction contains the rejection reason and the required tool name but no hidden answer, and a second violation fails the attempt. Model-facing handoffs contain role content only, while raw memory-record IDs, owner IDs, and evaluator fields remain inside the workflow adapter.

\subsection{Protected-action and replan harness}

\planfence{} follows the four-stage path below in both live workflows and controlled replay. Replay substitutes a recorded decision for the model call while retaining the validation and transport operations. The centralized replay control also rechecks its input after replanning. In the single-revision live comparison in Table~\ref{tab:rq1}(b), the centralized control instead checks before generation and then executes the revised action, and that setting contains no second update during generation. The overlapping-update comparison in Appendix~\ref{app:replan-overlap} requires a post-replan check for both policies.

\begin{tcolorbox}[
  enhanced,breakable,colback=green!3,colframe=teal!70!black,
  title=\textbf{Single-replan protected-action path},fonttitle=\bfseries,
  coltitle=white,colbacktitle=teal!70!black,boxrule=0.6pt,arc=1.5mm,
  left=1.5mm,right=1.5mm,top=2mm,bottom=1.5mm]
\small
\textbf{1. Validate.} Capture the locally available dependency IDs $C_a$, recover $F_a$ from the supplied plan roots, and query the owners for $H$. Release requires agreement among all three views.

\textbf{2. Refresh.} Fetch, verify, and install each owner record whose ID differs from $C_a$. If local memory is already current but the plan is not, request replanning without fetching that content again. Missing or invalid evidence blocks the action.

\textbf{3. Replan once.} The caller invokes the same role with isolated conversational history and a JSON object containing \texttt{replan\_required}, the old action, the fresh requirement, the fresh eligible candidates, and the public handoffs. It stores the replacement action with the exact refreshed requirement as its parent.

\textbf{4. Revalidate and propagate.} Capture the local dependency IDs again and validate the replacement root with the replan budget marked as used. On success, execute a single primary action and pass the new root and the validated dependency to the downstream auditor before its first model call. A remaining mismatch blocks execution.
\end{tcolorbox}

Using the recorded decision in controlled replay preserves the memory and network path while removing variation from an additional model call. The validator returns a status and the owner-reported record IDs, and it neither invokes the model nor returns the fetched payloads as part of that result. Successful fetches install the verified records in the local store, from which the caller obtains the content for replanning. The replacement root, and not the original root with a newer head substituted, supplies the evidence for the next validation. In the measured path, fetching changed content precedes the exhausted-budget decision, so a second mismatch can incur a fetch before the validator blocks.

\subsection{Exploratory live-workflow analysis}
\label{sec:live-appendix}

Before imposing the matched intervention, we examine whether model-generated workflows exhibit the fresh-memory, stale-plan pattern. Centralized-state competence controls establish that the model and tools can solve the workflow: 30 task pairs, each run under the compact and the large-artifact state tier (60 attempts), all complete, and all 480 role decisions satisfy their public tool schemas.

The pre-specified executor-state endpoint finds an obsolete local requirement in 3 of 30 runs, below the 15-of-30 threshold set in advance for that endpoint, which therefore fails. A post-hoc derivation endpoint, defined after inspecting these runs, counts runs in which the executor holds the revised requirement yet acts from a plan derived from the initial one. In 28 of 30 runs the plan cites the initial application requirement. Of these, 15 are the fresh-memory, stale-plan cases: the executor holds the revision yet issues the incorrect action from the stale plan; in 10 the executor holds the revision and issues the correct action despite the stale plan; and in 3 the executor's own copy is also obsolete and the action is incorrect (the three cases counted by the pre-specified endpoint). The remaining 2 runs have a current plan and a correct action. The post-hoc count coincides with the pre-specified 15-case number, and we do not treat it as having met a pre-registered criterion. The revision in these runs is a public workflow event generated and delivered by the harness and published by the owner role, whose tool schema permits exactly that event; the timing of its propagation relative to role calls is endogenous to the run. The 10 corrected actions establish behavior, not the model's reason for correcting. The 15 cases span deployment, fulfillment, and reservation with counts 6, 2, and 7. The model produces 240 accepted role decisions from 241 responses, with no unrecovered schema, workflow-contract, or infrastructure failure. Those plan records stored the application requirement ID in their public payload but did not attach the immutable memory record as a parent. This analysis establishes that generated plans can remain stale, and controlled replay and the matched live study provide the exact-parent comparison.

\subsection{Matched controlled-live check}

The matched study crosses three workflow families, 10 evaluation seeds, and three memory methods, for 90 attempts. Each task forms a plan over the initial requirement and then receives a single owner revision before execution. Owner-head freshness exposes the new record only after the executor's initial model call has generated its action, so that executor decides from obsolete input in all 30 attempts (as does \planfence{}'s executor before its validation step). \planfence{} and centralized lineage then detect the changed input and regenerate the action in a single isolated replanning call by the executor role. \planfence{} revalidates before acting, while the centralized live control executes the revised action without a second head retrieval. Both complete under this single-revision intervention, which does not test updates during generation.

Owner-head freshness issues the stale action and fails all 30 tasks. Centralized lineage and \planfence{} each complete 30 of 30 tasks with no invalid action and a single successful replan per task. Their auditors repeat 4 and 10 valid actions, respectively; no causal attribution for this difference is established here. All 90 attempts complete. Four of 784 model responses initially violate the tool schema and succeed after the permitted public correction, and no attempt ends in an infrastructure or tool-contract failure. Median role-call durations are 573, 591, and 576\,ms for centralized lineage, \planfence{}, and owner-head freshness; the replacement executor calls take 587 and 639\,ms at the median; task durations are 5.4, 5.7, and 4.9\,s; and the arms produce 273, 271, and 240 model responses with median token use of about 10.5k, 10.6k, and 9.2k per task.

% NEW-RESULTS-20260926 begin
\textbf{Fresh-before-generation control.} A later live campaign adds a control that refreshes the executor's inputs through the owner sockets before the executor generates, with a tool schema that permits both the inherited action and fresh locally eligible alternatives; it performs no derivation check. It ran as a separate campaign on two H100 GPUs with tensor parallelism 2, unlike the four-GPU original study, with nine disjoint smoke attempts followed by 90 main attempts (30 per arm) and no infrastructure failures. Table~\ref{tab:b4-live} reports the three arms. The fresh-before-generation executor adapts its primary action in 28 of 30 tasks (the two exceptions, fulfillment seeds 102 and 106, retain the old warehouse choice), yet completes none, because its downstream auditor keeps the original control path, reading local state and generating before the owner refresh, and issues an incorrect remedial action after the correct primary. \planfence{} completes 30 of 30, and its repair path hands the validated inputs to the auditor before that role's first call. What the cohort shows is that a model given current inputs adapts, and that adaptation at one role does not survive a stale role downstream; it varies executor input exposure only and does not test refreshing every role before generation, and we read 28 and 30 correct primaries as equivalent. Median task durations are 6.5, 5.6, and 5.6\,s and median token use 10.6k, 9.4k, and 9.3k for \planfence{}, the fresh-before-generation control, and original owner-head freshness, with 271, 240, and 240 model completions.

\begin{table}[ht]
\centering\small
\setlength{\tabcolsep}{6pt}
\begin{tabular}{lrr}
\toprule
Arm & Correct primary / scheduled & Task success / scheduled \\
\midrule
Owner-head freshness (original path) & 0/30 & 0/30 \\
Fresh-before-generation executor & 28/30 & 0/30 \\
\rowcolor{planfencefill} \planfence{} & 30/30 & 30/30 \\
\bottomrule
\end{tabular}
\caption{Live control that refreshes the executor's inputs before generation: a separate campaign with two GPUs and tensor parallelism 2, 30 tasks per arm.}
\label{tab:b4-live}
\end{table}
% NEW-RESULTS-20260926 end
 Each paired centralized-lineage and \planfence{} run follows the same workflow steps.

\subsection{Evaluation coverage}
\label{app:evidence-completeness}

\begin{figure}[t]
\centering
\includegraphics[width=0.90\textwidth]{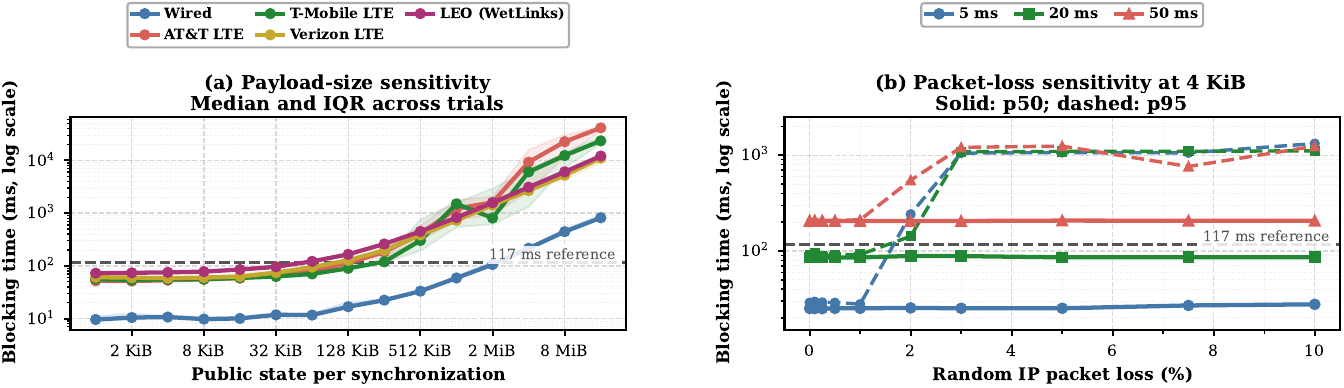}
\caption{Blocking time of a single fresh authenticated TCP transfer. (a) The time grows with public-state size under replayed historical LTE traces. (b) Packet loss inflates the tail (95th percentile) of a 4\,KiB transfer while the median remains stable. The 117\,ms reference equals 10\% of the median Qwen3.5 role-call duration in an earlier calibration (354 calls; median 1.17\,s, interquartile range 0.69--2.90\,s); role calls in the matched live study are faster (Appendix~\ref{app:implementation}). Only the loopback and LTE traces are used in the policy runs; the wired and LEO curves and the packet-loss panel characterize the transport and are not used in the policy comparisons.}
\label{fig:sync-boundary}
\end{figure}

Across the aligned 32{,}700-action policy grid (23{,}040 actions for the three primary policies across four networks and four update targets, 7{,}680 for the four delayed-sync variants on the AT\&T trace, 660 for the two freshness controls, and 1{,}320 for the four additional exact-lineage controls at $\rho=4$), the tested non-delayed validation controls remain derivation-current and available, while stale or delayed-sync controls account for all 4{,}143 invalid actions (3{,}483 delayed-sync and 660 freshness). The matched keyspace study completes 13{,}200 of 13{,}200 actions and supplies the scope-cost comparison. The symmetric artifact control completes 15{,}360 of 15{,}360 actions with zero violations of \eqref{eq:current-input}; its 960 episodes comprise 480 compact-state and 480 large-artifact episodes (7{,}680 actions each), and every artifact-tier action fetches and verifies the changed artifact twice. Its distinct action path tests payload integrity, while the compact-state grid supplies comparative stall. Figure~\ref{fig:sync-boundary} characterizes the transport used throughout: the blocking time of a single fresh authenticated TCP transfer grows with payload size under the replayed LTE traces, and packet loss inflates its tail but not its median.

\section{Controlled-Replay Ablations}
\label{app:controlled-replay}

The policy boundary is not confined to a single configuration. The high-update ordering persists across synchronization cadence, team size, and episode length, while the benefit of dependency scope narrows as an action approaches the full keyspace. A ``valid'' action is an issued action whose plan reaches every current declared input, and a blocked action never enters that numerator. Unless noted otherwise, the tables use compact records, the AT\&T trace, and $\rho=4$.

\begin{table}[t]
\centering
\footnotesize
\setlength{\tabcolsep}{3.0pt}
\begin{tabular}{llrrrr}
\toprule
Category & Method / variant & \shortstack{Invalid\\/ issued} & \shortstack{Available\\/ scheduled} & \shortstack{Stall\\(ms/action)} & \shortstack{Traffic\\(KiB/action)} \\
\midrule
\multirow{2}{*}{\shortstack[l]{Unsafe\\freshness}} & Local replica & 330/330 & 330/330 & 0.0 & 3.8 \\
 & Owner-head freshness & 330/330 & 330/330 & 151.8 & 7.1 \\
\addlinespace
\multirow{4}{*}{\shortstack[l]{Delayed sync\\(unsafe)}} & Metadata sync, $K=2$ & 42/330 & 330/330 & 333.1 & 23.1 \\
 & $\hookrightarrow\ K=4$ & 108/330 & 330/330 & 213.7 & 21.2 \\
 & $\hookrightarrow\ K=8$ & 222/330 & 330/330 & 122.4 & 17.5 \\
 & $\hookrightarrow\ K=16$ & 291/330 & 330/330 & 60.7 & 13.2 \\
\midrule
\multirow{7}{*}{\shortstack[l]{Derivation-\\current}} & Centralized lineage & 0/330 & 330/330 & 508.6 & \underline{15.6} \\
 & Metadata sync, $K=1$ & 0/330 & 330/330 & 403.4 & 23.5 \\
 & Majority-replica validation & 0/330 & 330/330 & 1075.3 & 223.7 \\
 & Per-key all-key validation & 0/330 & 330/330 & 282.8 & 19.2 \\
 & Batched all-key validation & 0/330 & 330/330 & \underline{258.4} & 16.4 \\
 & All-replica dependency validation & 0/330 & 330/330 & 342.2 & 56.0 \\
\rowcolor{planfencefill} & \textbf{\planfence{} (ours)} & 0/330 & 330/330 & \textbf{230.8} & \textbf{8.1} \\
\bottomrule
\end{tabular}
\caption{Complete primary replay controls at nominal $\rho=4$ (42 updates, 11 actions per episode), eight keys, and the AT\&T trace with new connections. Counts distinguish invalid issued actions from availability. Bold and underlined values identify the best and second-best point estimates among zero-invalid, fully available policies.}
\label{tab:all-controls}
\end{table}

\begin{figure}[t]
\centering
\includegraphics[width=0.97\textwidth]{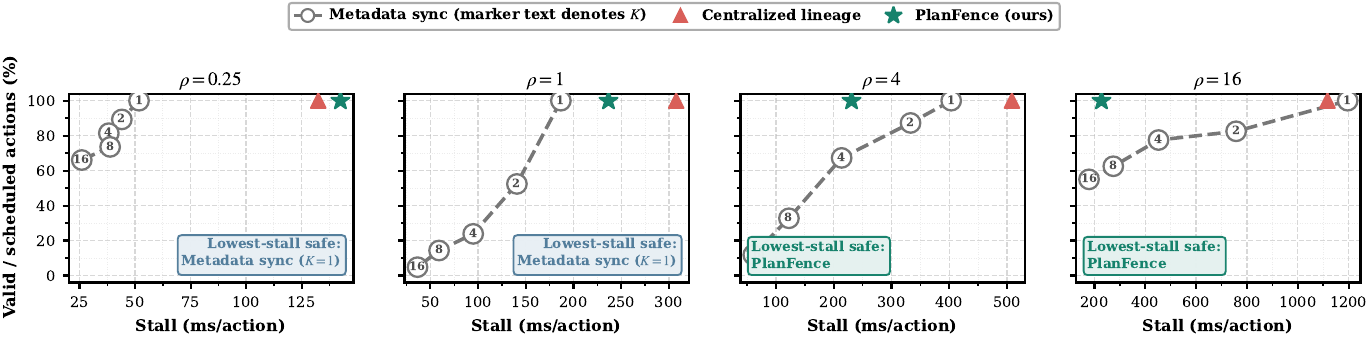}
\caption{Validity against stall on the AT\&T trace at four nominal update targets. Numbers inside the metadata-sync markers denote the interval $K$. Metadata sync ($K=1$) has lower stall at $\rho\leq1$, \planfence{} has lower stall at $\rho\geq4$, and $K>1$ reduces stall only by issuing invalid actions.}
\label{fig:cadence-tradeoff}
\end{figure}

Delaying metadata synchronization trades safety for lower stall. Figure~\ref{fig:cadence-tradeoff} shows this trade-off on the AT\&T trace, and Table~\ref{tab:cadence-appendix} reports the underlying counts and costs. Increasing $K$ reduces proactive coordination but leaves the announced head obsolete between barriers and admits more invalid actions. Tables~\ref{tab:team-appendix} and~\ref{tab:horizon-appendix} vary team size and episode length; the non-primary settings use one template per workflow family, so their medians are the middle template and serve as sensitivity checks rather than estimates. Across these $\rho=4$ checks, \planfence{} has lower stall than centralized lineage and metadata sync ($K=1$) for $N=3,5,8$ and $T=16,64,256$, with 8.0 to 8.2\,KiB of traffic per action. The observed team-size ordering is consistent with the fan-out term in the first-order model, without fitting that model to stall.

\begin{table*}[ht]
\centering\scriptsize
\begin{tabular}{llrrr}
\toprule
Setting & Method & Valid / scheduled & Stall (ms/action) & Traffic (KiB/action) \\
\midrule
$\rho=0.25$ & Metadata sync, $K=1$ & 840/840 & 51.8 & 5.8 \\
$\rho=0.25$ & $\hookrightarrow\ K=2$ & 750/840 & 44.1 & 5.5 \\
$\rho=0.25$ & $\hookrightarrow\ K=4$ & 684/840 & 38.2 & 5.4 \\
$\rho=0.25$ & $\hookrightarrow\ K=8$ & 618/840 & 38.8 & 5.3 \\
$\rho=0.25$ & $\hookrightarrow\ K=16$ & 555/840 & 26.0 & 5.1 \\
\rowcolor{planfencefill} $\rho=0.25$ & \textbf{\planfence{} (ours)} & 840/840 & 142.5 & 5.8 \\
\addlinespace
$\rho=1$ & Metadata sync, $K=1$ & 630/630 & 186.8 & 11.6 \\
$\rho=1$ & $\hookrightarrow\ K=2$ & 330/630 & 140.7 & 10.6 \\
$\rho=1$ & $\hookrightarrow\ K=4$ & 150/630 & 94.9 & 9.7 \\
$\rho=1$ & $\hookrightarrow\ K=8$ & 90/630 & 59.1 & 9.0 \\
$\rho=1$ & $\hookrightarrow\ K=16$ & 30/630 & 36.5 & 7.6 \\
\rowcolor{planfencefill} $\rho=1$ & \textbf{\planfence{} (ours)} & 630/630 & 236.7 & 8.1 \\
\addlinespace
$\rho=4$ & Metadata sync, $K=1$ & 330/330 & 403.4 & 23.5 \\
$\rho=4$ & $\hookrightarrow\ K=2$ & 288/330 & 333.1 & 23.1 \\
$\rho=4$ & $\hookrightarrow\ K=4$ & 222/330 & 213.7 & 21.2 \\
$\rho=4$ & $\hookrightarrow\ K=8$ & 108/330 & 122.4 & 17.5 \\
$\rho=4$ & $\hookrightarrow\ K=16$ & 39/330 & 60.7 & 13.2 \\
\rowcolor{planfencefill} $\rho=4$ & \textbf{\planfence{} (ours)} & 330/330 & 230.8 & 8.1 \\
\addlinespace
$\rho=16$ & Metadata sync, $K=1$ & 120/120 & 1196.3 & 65.8 \\
$\rho=16$ & $\hookrightarrow\ K=2$ & 99/120 & 757.2 & 61.5 \\
$\rho=16$ & $\hookrightarrow\ K=4$ & 93/120 & 451.9 & 53.7 \\
$\rho=16$ & $\hookrightarrow\ K=8$ & 75/120 & 273.9 & 42.6 \\
$\rho=16$ & $\hookrightarrow\ K=16$ & 66/120 & 178.4 & 29.6 \\
\rowcolor{planfencefill} $\rho=16$ & \textbf{\planfence{} (ours)} & 120/120 & 227.3 & 8.1 \\
\bottomrule
\end{tabular}
\caption{Synchronization cadence across nominal update targets. Waiting more than a single work unit reduces proactive cost by allowing invalid stale-plan actions. Centralized lineage at $\rho\in\{0.25,1,16\}$ was run in the same sweep and is plotted in Figure~\ref{fig:cadence-tradeoff}; it is the runner-up in the corresponding AT\&T cells of Figure~\ref{fig:policy-boundary}(a), and its medians are not tabulated here.}
\label{tab:cadence-appendix}
\end{table*}

\begin{table*}[ht]
\centering\scriptsize
\begin{tabular}{llrrr}
\toprule
Setting & Method & Valid / scheduled & Stall (ms/action) & Traffic (KiB/action) \\
\midrule
$N=3$ & Centralized lineage & 33/33 & 530.2 & 15.9 \\
$N=3$ & Metadata sync, $K=1$ & 33/33 & 381.1 & 15.7 \\
$N=3$ & Per-key all-key validation & 33/33 & 290.4 & 16.8 \\
$N=3$ & Batched all-key validation & 33/33 & 237.0 & 12.8 \\
$N=3$ & All-replica dependency validation & 33/33 & 300.5 & 36.7 \\
$N=3$ & Majority-replica validation & 33/33 & 953.4 & 138.8 \\
\rowcolor{planfencefill} $N=3$ & \textbf{\planfence{} (ours)} & 33/33 & 223.5 & 8.0 \\
\addlinespace
$N=5$ & Centralized lineage & 330/330 & 508.6 & 15.6 \\
$N=5$ & Metadata sync, $K=1$ & 330/330 & 403.4 & 23.5 \\
$N=5$ & Per-key all-key validation & 330/330 & 282.8 & 19.2 \\
$N=5$ & Batched all-key validation & 330/330 & 258.4 & 16.4 \\
$N=5$ & All-replica dependency validation & 330/330 & 342.2 & 56.0 \\
$N=5$ & Majority-replica validation & 330/330 & 1075.3 & 223.7 \\
\rowcolor{planfencefill} $N=5$ & \textbf{\planfence{} (ours)} & 330/330 & 230.8 & 8.1 \\
\addlinespace
$N=8$ & Centralized lineage & 33/33 & 596.3 & 16.1 \\
$N=8$ & Metadata sync, $K=1$ & 33/33 & 489.1 & 36.8 \\
$N=8$ & Per-key all-key validation & 33/33 & 292.9 & 20.6 \\
$N=8$ & Batched all-key validation & 33/33 & 301.9 & 21.0 \\
$N=8$ & All-replica dependency validation & 33/33 & 383.6 & 95.5 \\
$N=8$ & Majority-replica validation & 33/33 & 1195.3 & 501.1 \\
\rowcolor{planfencefill} $N=8$ & \textbf{\planfence{} (ours)} & 33/33 & 235.1 & 8.2 \\
\bottomrule
\end{tabular}
\caption{Team-size sensitivity with a single dependency per action. Rows other than $N=5$ use one template per workflow family (33 scheduled actions), so their medians are the middle template and are sensitivity checks rather than estimates. The ordering \planfence{} $<$ metadata sync $<$ centralized lineage holds in every row; the ordering among the remaining controls does not (per-key and batched all-key validation swap at $N=8$, and \planfence{} versus batched all-key is within 10\% at $N=3$).}
\label{tab:team-appendix}
\end{table*}

\begin{table*}[ht]
\centering\scriptsize
\begin{tabular}{llrrr}
\toprule
Setting & Method & Valid / scheduled & Stall (ms/action) & Traffic (KiB/action) \\
\midrule
$T=16$ & Centralized lineage & 9/9 & 452.4 & 15.0 \\
$T=16$ & Metadata sync, $K=1$ & 9/9 & 347.6 & 21.4 \\
\rowcolor{planfencefill} $T=16$ & \textbf{\planfence{} (ours)} & 9/9 & 210.6 & 8.1 \\
\addlinespace
$T=64$ & Centralized lineage & 330/330 & 508.6 & 15.6 \\
$T=64$ & Metadata sync, $K=1$ & 330/330 & 403.4 & 23.5 \\
\rowcolor{planfencefill} $T=64$ & \textbf{\planfence{} (ours)} & 330/330 & 230.8 & 8.1 \\
\addlinespace
$T=256$ & Centralized lineage & 129/129 & 593.4 & 16.3 \\
$T=256$ & Metadata sync, $K=1$ & 129/129 & 523.0 & 24.7 \\
\rowcolor{planfencefill} $T=256$ & \textbf{\planfence{} (ours)} & 129/129 & 286.0 & 8.1 \\
\bottomrule
\end{tabular}
\caption{Episode-length sensitivity for the three primary policies. Rows other than $T=64$ use one template per workflow family (9 and 129 scheduled actions; the schedule generator yields 3, 11, and 43 protected actions per episode at $T=16$, 64, and 256). The ordering persists from 16 to 256 work units in this three-template check.}
\label{tab:horizon-appendix}
\end{table*}

\begin{table*}[ht]
\centering\scriptsize
\begin{tabular}{llrrrr}
\toprule
Network & Method & Keys & Valid / scheduled & Stall (ms/action) & Traffic (KiB/action) \\
\midrule
Loopback & Batched all-key validation & 8 & 330/330 & 115.5 & 16.4 \\
Loopback & \planfence{} (ours) & 8 & 330/330 & 67.8 & 8.1 \\
Loopback & Batched all-key validation & 16 & 330/330 & 127.7 & 21.4 \\
Loopback & \planfence{} (ours) & 16 & 330/330 & 67.1 & 8.1 \\
Loopback & Batched all-key validation & 32 & 330/330 & 131.4 & 30.6 \\
Loopback & \planfence{} (ours) & 32 & 330/330 & 67.0 & 8.1 \\
Loopback & Batched all-key validation & 64 & 330/330 & 154.6 & 47.8 \\
Loopback & \planfence{} (ours) & 64 & 330/330 & 64.5 & 8.1 \\
Loopback & Batched all-key validation & 128 & 330/330 & 199.2 & 81.7 \\
Loopback & \planfence{} (ours) & 128 & 330/330 & 66.0 & 8.1 \\
\addlinespace
AT\&T trace & Batched all-key validation & 8 & 990/990 & 365.7 & 16.4 \\
AT\&T trace & \planfence{} (ours) & 8 & 990/990 & 295.3 & 8.1 \\
AT\&T trace & Batched all-key validation & 16 & 990/990 & 400.1 & 21.4 \\
AT\&T trace & \planfence{} (ours) & 16 & 990/990 & 312.0 & 8.1 \\
AT\&T trace & Batched all-key validation & 32 & 990/990 & 419.4 & 30.6 \\
AT\&T trace & \planfence{} (ours) & 32 & 990/990 & 300.3 & 8.1 \\
AT\&T trace & Batched all-key validation & 64 & 990/990 & 449.3 & 47.8 \\
AT\&T trace & \planfence{} (ours) & 64 & 990/990 & 282.4 & 8.1 \\
AT\&T trace & Batched all-key validation & 128 & 990/990 & 428.7 & 81.7 \\
AT\&T trace & \planfence{} (ours) & 128 & 990/990 & 345.2 & 8.1 \\
\bottomrule
\end{tabular}
\caption{Complete aligned semantic-keyspace comparison. Loopback rows aggregate 30 templates, and AT\&T rows aggregate the same templates at three independent trace offsets. Both methods use exact parent binding, a single post-replan validation, and a new TCP connection per RPC.}
\label{tab:keyspace-appendix}
\end{table*}

In Table~\ref{tab:keyspace-appendix}, all-key traffic rises by about 0.54\,KiB per action per additional key between eight and 128 keys. This measured per-action increment is consistent with the unrelated-entry term in \eqref{eq:scope}; it is not an estimate of the per-entry, per-pass $c_e$ without reconciling encoding and pass counts. Scoped traffic remains flat in this fixed single-dependency configuration.

\runinhead{Large-artifact control.} This control crosses 30 templates, four policies, loopback and the AT\&T trace, two state tiers (compact records and 256--320\,KiB artifacts), and nominal update targets $\rho\in\{1,4\}$ (21 and 11 protected actions per episode) at $N=5$, $T=64$, $d=1$: 960 episodes and 15{,}360 protected actions, half in each tier. In the artifact tier every policy fetches and verifies the same changed public artifact twice per action; compact episodes fetch no artifacts. Every exact-parent action satisfies \eqref{eq:current-input}. This control verifies the payload path, while the matched compact-state grid supplies comparative stall.

\begin{table}[ht]
\centering
\footnotesize
\setlength{\tabcolsep}{5pt}
\begin{tabular}{lrrrrr}
\toprule
& & \multicolumn{2}{c}{Stall (ms/action)} & \multicolumn{2}{c}{Traffic (KiB/action)} \\
\cmidrule(lr){3-4}\cmidrule(lr){5-6}
Dependencies & Valid / scheduled & Batched all-key & \planfence{} & Batched all-key & \planfence{} \\
\midrule
$d=1$ & 330/330 & 262.8 & \textbf{223.6} & 16.4 & \textbf{8.1} \\
$d=2$ & 330/330 & 270.7 & 245.2 & 19.5 & \textbf{13.2} \\
$d=8=|\mathcal X|$ & 330/330 & 400.1 & 421.7 & 34.1 & 36.8 \\
\bottomrule
\end{tabular}
\caption{Widening the declared dependency set at a fixed 8-key state and $\rho=4$ on the AT\&T trace, over 30 templates under the same one-replan rule, in a separate campaign. Both methods are valid on every scheduled action; bold marks a cost at least 10\% lower than the other method, and unmarked pairs are ties.}
\label{tab:fanin}
\end{table}

% NEW-RESULTS-20260926 begin
\subsection{Contemporaneous four-policy campaign and headline replication}
\label{app:a2-a1}
Two later campaigns on the current harness revision re-run the timing comparison with all policies executed together, so that no comparison mixes campaigns. Both use the 30 templates, $N=5$, $T=64$, $d=1$, eight keys, the AT\&T trace at offset zero, and a new TCP connection per RPC unless stated. Table~\ref{tab:a2-campaign} runs metadata sync ($K=1$), centralized lineage, batched all-key validation, and \planfence{} at all four update targets (480 episodes). Metadata sync has the lowest median stall at $\rho\le1$ and \planfence{} at $\rho\ge4$, as in the primary sweep, and \planfence{}'s stall is 19.7--24.9\% below batched all-key validation at every target; across the 64$\times$ range of $\rho$, \planfence{} varies by 1.8$\times$ and metadata sync by 21$\times$. Table~\ref{tab:a1-replication} repeats the $\rho=4$ headline configuration in three independent fresh-connection batches and one persistent-connection batch (five policies including per-key all-key validation; 600 episodes): the \planfence{} to batched all-key stall reduction is 23.7--24.1\% in every fresh batch and 15.1\% with reuse. Across both campaigns all 1{,}080 attempts are accounted for; one metadata-sync episode and two per-key all-key episodes abort and their 33 scheduled slots receive no imputed outcome or latency; ten further actions are blocked within completed episodes; availability is 14{,}237 of 14{,}280 with zero recorded invalid actions, and \planfence{} completes 3{,}240 of 3{,}240. The primary campaign of Table~\ref{tab:causal-headline} ran on an earlier harness revision; Appendix~\ref{app:statistics} explains why we compare orderings, not absolute stall, across revisions.

\begin{table}[ht]
\centering\small
\setlength{\tabcolsep}{5pt}
\begin{tabular}{lrrrrr}
\toprule
$\rho$ & Metadata sync, $K=1$ & Centralized lineage & Batched all-key & \planfence{} & \planfence{} vs.\ batched \\
\midrule
0.25 & 68.3 & 150.7 & 235.2 & 176.6 & $-24.9\%$ \\
1 & 229.5 & 358.2 & 387.2 & 310.9 & $-19.7\%$ \\
4 & 510.7 & 629.5 & 342.2 & 257.3 & $-24.8\%$ \\
16 & 1411.1 & 1233.6 & 341.1 & 264.8 & $-22.4\%$ \\
\bottomrule
\end{tabular}
\caption{Median stall (ms/action) in the contemporaneous four-policy campaign on the AT\&T trace at offset zero with a new connection per RPC, 30 templates per cell. The last column is \planfence{}'s reduction relative to batched all-key validation.}
\label{tab:a2-campaign}
\end{table}

\begin{table}[ht]
\centering\small
\setlength{\tabcolsep}{6pt}
\begin{tabular}{llrrr}
\toprule
Batch & Connections & \planfence{} & Batched all-key & Reduction \\
\midrule
1 & new per RPC & 255.4 & 336.5 & 24.1\% \\
1 & persistent & 226.6 & 266.9 & 15.1\% \\
2 & new per RPC & 250.0 & 328.9 & 24.0\% \\
3 & new per RPC & 257.3 & 337.1 & 23.7\% \\
\bottomrule
\end{tabular}
\caption{Headline configuration ($\rho=4$, eight keys, AT\&T offset zero) replicated on the current harness revision; stall in ms/action, medians over 30 templates. Batched all-key validation has 328 of 330 available actions in the persistent-connection batch, and two per-key all-key episodes abort in batch 3.}
\label{tab:a1-replication}
\end{table}

% NEW-RESULTS-20260926 end
\subsection{Independent AT\&T trace phases}

Changing the start point within the AT\&T trace changes absolute latency but preserves the pointwise ordering. Table~\ref{tab:additional-controls} reports the independently replayed 0, 5, and 15 second offsets at both keyspace endpoints. Both exact-parent methods complete every action, and \planfence{} has lower median stall and traffic in all 6 paired aggregates. The paired stall interval excludes zero in five of the six 30-template strata. At eight keys and the 5-second offset it includes zero, and the point-estimate difference also falls within the 10\% practical-tie threshold.

\begin{table*}[ht]
\centering\small
\setlength{\tabcolsep}{5pt}
\renewcommand{\arraystretch}{0.96}
\begin{tabular}{llrrrr}
\toprule
Keys & Trace offset & Method & Valid / scheduled & Stall & Traffic \\
 & & & & (ms/action) & (KiB/action) \\
\midrule
8 & 0 s & Batched all-key validation & 330/330 & 264.5 & 16.4 \\
\rowcolor{planfencefill} 8 & 0 s & \textbf{\planfence{} (ours)} & 330/330 & \textbf{223.9} & \textbf{8.1} \\
8 & 5 s & Batched all-key validation & 330/330 & 365.7 & 16.4 \\
\rowcolor{planfencefill} 8 & 5 s & \textbf{\planfence{} (ours)} & 330/330 & 348.4 & \textbf{8.1} \\
8 & 15 s & Batched all-key validation & 330/330 & 390.3 & 16.4 \\
\rowcolor{planfencefill} 8 & 15 s & \textbf{\planfence{} (ours)} & 330/330 & \textbf{295.3} & \textbf{8.1} \\
\addlinespace
128 & 0 s & Batched all-key validation & 330/330 & 350.0 & 81.7 \\
\rowcolor{planfencefill} 128 & 0 s & \textbf{\planfence{} (ours)} & 330/330 & \textbf{239.4} & \textbf{8.1} \\
128 & 5 s & Batched all-key validation & 330/330 & 428.7 & 81.7 \\
\rowcolor{planfencefill} 128 & 5 s & \textbf{\planfence{} (ours)} & 330/330 & \textbf{345.2} & \textbf{8.1} \\
128 & 15 s & Batched all-key validation & 330/330 & 643.3 & 81.7 \\
\rowcolor{planfencefill} 128 & 15 s & \textbf{\planfence{} (ours)} & 330/330 & \textbf{390.8} & \textbf{8.1} \\
\bottomrule
\end{tabular}
\caption{Independent AT\&T trace phases under the aligned comparison. Both methods fetch a changed record once, replan once, and validate again. Bold marks a cost at least 10\% lower within each offset and keyspace pair; the 8-key, 5\,s stall pair is a tie. These are campaigns separate from Table~\ref{tab:causal-headline}.}
\label{tab:additional-controls}
\end{table*}

\clearpage
\subsection{Learned policy selection does not dominate a transparent rule under distribution shift}

Could a learned selector reduce coordination cost without weakening the safety path? We allow it to choose among centralized lineage, metadata sync ($K=1$), and \planfence{}, all of which retain the deterministic derivation check, and malformed or unavailable selector output falls back to \planfence{}. The transparent reference uses metadata sync if $\rho\leq1$ and \planfence{} otherwise, which summarizes the observed low/high-update ordering; learned selectors may also choose centralized lineage. This empirical reference rule is not obtained by calibrating \eqref{eq:crossover}.

The study derives 1{,}332 examples from controlled replay, each aggregating 6 workflows from three families. A policy is \emph{eligible} if it issues no invalid action and completes every scheduled action. Among eligible policies within 10\% of the lowest stall, the target is the policy with the least wire traffic, and we term this a \emph{safe near-optimal} choice. The split contains 648 training, 144 development, and 180 in-distribution (ID) examples. Four disjoint 90-example out-of-distribution (OOD) blocks hold out, respectively, the Verizon trace, interpolated update rates $\{0.5,2,8\}$, endpoint key counts $\{8,128\}$, and their compound shift. Selectors see only pre-action workload dimensions and static trace calibration, and never profile identity, workflow text, outcomes, or future trace events. Neural results use three-seed ensembles.

Learned selectors improve ID selection but do not consistently dominate the transparent rule under shift. Table~\ref{tab:safe-router} reports 95.0 to 96.7\% safe near-optimal selection for the learned models on ID examples, compared with 89.4\% for the transparent rule. Under shift, the tree and tabular networks fall to 84.4 to 94.2\%. DeBERTa-v3-base reaches 96.1\%, but its three-model inference estimate is 34.08\,ms, computed by summing three separate batch-one p95 measurements. We admit a learned selector only if every held-out block preserves completion and improves at least a single selection or cost criterion without worsening the others after inference. The tabular models improve only the keyspace block, while neither DeBERTa model yields an across-block improvement once inference cost is included. The derivation check remains deterministic under every selector.

\begin{table*}[ht]
\centering\small
\setlength{\tabcolsep}{3.2pt}
\renewcommand{\arraystretch}{0.96}
\begin{tabular}{lrrrrrc}
\toprule
& \multicolumn{2}{c}{Safe near-optimal $\uparrow$ (\%)}
& \multicolumn{2}{c}{Tail stall regret $\downarrow$ (ms/action)}
& Tail inference $\downarrow$ & All-shift \\
\cmidrule(lr){2-3}\cmidrule(lr){4-5}
Selector & ID & OOD & ID & OOD & (ms/query) & dominance \\
\midrule
Transparent rate rule & 89.4 & 95.0 & 24.7 & 26.5 & -- & reference \\
Fixed \planfence{} & 72.2 & 68.6 & 134.8 & 98.9 & -- & -- \\
\midrule
Depth-3 decision tree & 95.0 & 84.4 & 17.6 & 75.7 & 0.08 & no \\
Two-layer MLP & 96.7 & 94.2 & 7.5 & 35.4 & 0.21 & no \\
Residual MLP & 96.1 & 91.4 & 11.3 & 46.6 & 0.52 & no \\
DeBERTa-v3-base & 95.0 & 96.1 & 17.6 & 21.7 & 34.08 & no \\
DeBERTa-v3-large & 95.0 & 94.7 & 17.6 & 31.4 & 64.06 & no \\
\bottomrule
\end{tabular}
\caption{Selection among policies that retain the deterministic derivation check. OOD pools four disjoint 90-example shifts. Stall-regret tail values are 95th percentiles relative to the fastest eligible policy. Inference is reported separately and included in the blockwise dominance criterion. Its ensemble estimate sums three separate batch-one p95 measurements; this is not a measured ensemble p95. ``No'' means that the learned selector does not dominate the transparent rule in every OOD block.}
\label{tab:safe-router}
\end{table*}

\section{Validation Assumptions and Dependency Scope}
\label{app:validation-assumptions}

\subsection{Validation procedure}
Algorithm~\ref{alg:planfence} is the validation procedure of Section~\ref{sec:method}: it reconstructs the dependency frontier from the plan roots, queries the owners of the declared keys, refreshes any changed input, and then releases the action, requests the single permitted replan, or blocks.
\begin{algorithm}[!ht]
\small
\DontPrintSemicolon
\KwIn{action $a$, roots $L(a)$, declared keys $D(a)$, captured local IDs $C_a$,
      flag \texttt{replanned} (initially false)}
\lIf{roots, captured inputs, or required owner mappings are unavailable}{\Return \texttt{blocked}}
\tcp{Recover the plan's inputs, stopping at declared keys}
$Q\leftarrow L(a)$; $V\leftarrow\varnothing$; $B(x)\leftarrow\varnothing$ for $x\in D(a)$\;
\While{$Q$ is not empty}{
    Remove a record ID $i$ from $Q$\;
    \lIf{$i\in V$}{\texttt{continue}}
    Retrieve immutable record $r$ with ID $i$\;
    \lIf{$r$ is unavailable}{\Return \texttt{blocked}}
    $V\leftarrow V\cup\{i\}$\;
    \eIf{$\operatorname{key}(r)\in D(a)$}{
        $B(\operatorname{key}(r))\leftarrow B(\operatorname{key}(r))\cup\{i\}$\;
    }{
        Add the exact parent IDs of $r$ to $Q$\;
    }
}
\lIf{some $x\in D(a)$ has $|B(x)|\ne1$}{\Return \texttt{blocked}}
$F_a(x)\leftarrow$ the unique ID in $B(x)$, for each $x\in D(a)$\;
\tcp{Check owner heads and obtain changed content}
$H(x)\leftarrow$ concurrently query $o(x)$ for each $x\in D(a)$\;
\lIf{an owner query or response check fails}{\Return \texttt{blocked}}
Fetch, verify, and install records $H(x)$ where $H(x)\ne C_a(x)$\;
\lIf{a required fetch or verification fails}{\Return \texttt{blocked}}
\tcp{Request a single replacement plan, or release the action}
\If{$H\ne C_a$ or $F_a\ne C_a$}{
    \lIf{\texttt{replanned}}{\Return \texttt{blocked}}
    \Return \texttt{replan-required} with owner-reported IDs $H$\;
}
\Return \texttt{release}\;
\caption{Recovering a plan's inputs and validating a protected action.}
\label{alg:planfence}
\end{algorithm}

\subsection{Cost-model detail}
The model holds $\nu$ fixed. In an actual workload, unchanged candidates may need one pass, successful replanning may need two, and early failures may avoid queries. If update frequency changes replanning probability, it can change $\nu$ and therefore scoped work.

For matched pass multiplicity, the difference between all-key and scoped work in \eqref{eq:work} is
\begin{equation}
W_{\rm all}-W_{\rm scoped}\approx
\nu\bigl[(M-d)c_r+(|\mathcal X|-d)c_e\bigr].
\label{eq:scope}
\end{equation}
At fixed $M,d,\nu$ and message format, the unrelated-entry term grows linearly with key count; adding owners changes $M$ as well. The difference need not be positive, because batching can save more request overhead than scoping saves entries, and at full scope the unrelated-entry benefit is zero while the request term $\nu(M-|\mathcal X|)c_r\leq0$ remains.

\subsection{Assumptions and scope}

\planfence{} compares the recorded dependency frontier with the versions returned by the corresponding owners during validation. This check assumes benign owners, authenticated responses, immutable records and parent links, and a complete $D(a)$. Responses from different owners need not describe a common snapshot, and passing validation does not prevent an input from changing before the external effect. Byzantine ownership, owner migration, inferred dependencies, semantic merging, and private reasoning are outside the evaluated contract. Applications requiring cross-owner consistency or atomic check-and-execute behavior need additional coordination beyond the validator.

\textbf{Traversal and boundary coverage.} The worklist starts with the distinct IDs in $L(a)$, and every record is expanded at most once. A record whose key is outside $D(a)$ contributes its exact parent IDs to the worklist. A record whose key is in $D(a)$ contributes its own ID to that key's boundary set and is not expanded further. The traversal processes the remaining worklist even after finding a version for every key, since another branch may contain a conflicting version. Missing records encountered before or at a boundary block validation, and older records behind a reached boundary need not be present. A key lying only behind another declared boundary is not reached by this procedure and therefore fails coverage unless another branch reaches it, as required by the boundary-coverage clause of Assumption~\ref{as:all}.

For each $x\in D(a)$, $B(x)$ contains the distinct record IDs reached at that key before crossing any declared boundary. Shared branches or multiple roots reaching the same ID do not create a conflict. The frontier is defined by $F_a(x)=i$ only if $B(x)=\{i\}$ for every declared key. An empty set means that an input is missing, and a set with more than a single ID means that its version is ambiguous. Both cases block before owner queries. This set-valued construction explains why finding a single current-looking ancestor is insufficient, since it cannot erase a different input version reached along another branch.

Omitting a true input leaves its changes unchecked. Adding a key is conservative only if the plan also carries that key's exact input version, and an arbitrary extra key without such evidence is not a free safety improvement. Batched all-key validation uses a different operation. It queries every public head but compares only the action's actual inputs, so unrelated queried keys need not appear in the plan's ancestry, and its larger query scope should not be confused with expanding $D(a)$.

\textbf{Local capture and owner validation.} Before invoking the validator, the wrapper retrieves each declared key from local memory and captures its available record ID as $C_a(x)$, rejecting unavailable inputs or a record authored by a different owner. The validator separately reconstructs $F_a$ from the supplied roots and obtains $H$ through concurrent per-key owner queries. Responses are checked for the expected key and owner, valid record-ID and metadata fields, and conflicting owner sequences, and a required query failure blocks execution. For every $H(x)\ne C_a(x)$, the executor fetches that exact ID from the declared owner, verifies the content digest and author identity, and installs the record. These operations do not replace missing ancestry with a current owner head. Because the traversal reads a boundary record to learn its key, superseded versions must remain resolvable while plans derived from them are live; we do not evaluate retention policies.

The implemented release predicate is $F_a=C_a=H$, which implies \eqref{eq:current-input}. The intermediate local view also determines refresh and recovery behavior. If $F_a\ne C_a=H$, the local record is current but the plan needs replacement. If $C_a\ne H$, the validator refreshes the local view and requests a replacement, including the case in which $F_a=H$ already holds, which is the second construction in Appendix~\ref{app:proofs}. Thus the implementation can request replanning more conservatively than \eqref{eq:current-input} alone requires. After a single replacement, either mismatch blocks, and successful refreshes still occur before that terminal decision, as specified in Appendix~\ref{app:implementation}.

\section{Updates During Replanning}
\label{app:replan-overlap}

A fresh plan can become outdated before generation finishes. To examine what a single bounded replan can accomplish under continuing changes, we replay the 30 workflow templates with an independent owner updating the requirement. The executor captures its replan inputs before a synthetic delay $\delta\in\{0,1,2\}$ seconds. Additional relevant updates follow a seeded Poisson schedule at $\lambda\in\{0,0.25,1,2\}$ updates per second, paired across policies and delay settings. Each update receives a new record ID and alternates the public requirement between the two supported task states. The recorded decision is selected from the captured input and not from the head found after the delay. These delays are sensitivity settings and not measured model-generation times.

Both policies use a 30-second action deadline, a 10-second RPC timeout, at most a single replan, and a final version check. The updating owner continues through validation and the recorded action effect. \planfence{} uses five process-isolated replicas over authenticated loopback sockets, and centralized lineage uses the shared-state service of the live control. Unlike the central metadata directory in the timing study, this control stores the public content itself. An additional control changes only an unrelated key at two updates per second with a two-second delay.

Relevant changes, and not general background activity, exhaust the recovery budget. At a two-second delay and two relevant updates per second, both policies block all 30 actions, and with the same delay and only unrelated updates, both complete all 30. Intermediate rates produce the completion boundary in Table~\ref{tab:replan-overlap}. The policies need not encounter identical conflicts even with paired arrival schedules, because their retrievals and checks occur at different times. With no replanning delay \planfence{} blocks more often than the centralized control (1, 3 and 5 of 30 versus 0, 1 and 0 at 0.25, 1 and 2 updates/s), because its check spans more sequential owner exchanges; at one and two seconds of delay the two policies block within one trial of each other apart from initialization failures, so the completion boundary is set by the single-replan budget and the absence of atomicity rather than by where validation happens.

\begin{table}[!htbp]
\centering\small
\setlength{\tabcolsep}{4pt}
\begin{tabular}{@{}rrrrrrrrrr@{}}
\toprule
& & \multicolumn{4}{c}{Centralized lineage} & \multicolumn{4}{c}{\planfence{}} \\
\cmidrule(lr){3-6}\cmidrule(lr){7-10}
$\delta$ (s) & $\lambda$ (/s) & Complete & Block & Infra. & Gap & Complete & Block & Infra. & Gap \\
\midrule
0 & 0 & 30/30 & 0 & 0 & 0 & 30/30 & 0 & 0 & 0 \\
0 & 0.25 & 30/30 & 0 & 0 & 0 & 28/30 & 1 & 1 & 0 \\
0 & 1 & 29/30 & 1 & 0 & 0 & 27/30 & 3 & 0 & 0 \\
0 & 2 & 30/30 & 0 & 0 & 0 & 24/30 & 5 & 0 & 1 \\
\addlinespace
1 & 0 & 30/30 & 0 & 0 & 0 & 29/30 & 0 & 1 & 0 \\
1 & 0.25 & 23/30 & 7 & 0 & 0 & 22/30 & 7 & 1 & 0 \\
1 & 1 & 8/30 & 22 & 0 & 0 & 9/30 & 21 & 0 & 0 \\
1 & 2 & 4/30 & 26 & 0 & 0 & 4/30 & 26 & 0 & 0 \\
\addlinespace
2 & 0 & 30/30 & 0 & 0 & 0 & 28/30 & 0 & 2 & 0 \\
2 & 0.25 & 16/30 & 14 & 0 & 0 & 16/30 & 14 & 0 & 0 \\
2 & 1 & 4/30 & 26 & 0 & 0 & 4/30 & 25 & 1 & 0 \\
2 & 2 & 0/30 & 30 & 0 & 0 & 0/30 & 30 & 0 & 0 \\
\midrule
\multicolumn{10}{l}{Unrelated updates only, $\delta=2$ s, $\lambda_{\rm relevant}=0$, $\lambda_{\rm unrelated}=2$/s} \\
2 & 0 & 30/30 & 0 & 0 & 0 & 30/30 & 0 & 0 & 0 \\
\bottomrule
\end{tabular}
\caption{Completion under updates that continue during replanning, with 30 single-action trials per policy and row. Complete means an issued action that satisfies the current public requirement at the recorded effect point, Block counts policy refusals, Infra. counts initialization failures, and Gap counts inputs that changed after validation and before effect.}
\label{tab:replan-overlap}
\end{table}

The owner logs also distinguish validation correctness from effect-time currentness. All 264 issued centralized actions and 252 issued \planfence{} actions match the owner versions returned during their checks. A single \planfence{} input changes after the final head response but before the recorded effect, about 9\,ms after the owner head was delivered to the executor's replica, leaving 251 effect-current actions. This observation illustrates why \eqref{eq:current-input} is not an atomic check-and-execute guarantee. Six \planfence{} trials time out during initialization and remain in the scheduled denominators, and separately linked reruns of those 6 trials finish with unchanged settings, without replacing their original outcomes.

% NEW-RESULTS-20260926 begin
\textbf{Live cohort with an in-generation update.} A separate live cohort replaces the synthetic delay with a real one: for each of centralized lineage and \planfence{}, nine initial-only control attempts and nine attempts in which the owner commits a relevant update while the replacement generation is in flight (commit-inside-generation ordering confirmed from the event logs; 36 main and 12 smoke attempts). All nine controls per policy issue an effect-correct primary action, and all nine overlapped attempts per policy block after the single replan. The cohort is small (nine attempts per condition, no unrelated-update control); the synthetic-delay study above supplies the missing control and the larger counts.
% NEW-RESULTS-20260926 end

\FloatBarrier
\section{Native Validation of Inherited Plans}
\label{app:validator-comparison}

A later executor needs the versions behind the inherited plan and not only the versions it has recently observed. Section~\ref{sec:rq1} compares \planfence{} with the native single-node S-Bus service~\citep{khan2026sbus}, using its session-observation tracking and its declared version-set interface as separate configurations. The application adapter stores the public requirement and the recorded plan, publishes the requirement revision through the native commit API, and submits the proposed action through that API's validation path. The inherited version set is constructed from the plan's public input records, without grader information. Native session retention and validation remain enabled, and no observations are erased.

The scenarios in Table~\ref{tab:validator-comparison} separate unchanged inputs, changes after planning within a single session, and a handoff to a later executor that retrieves the new requirement while retaining the old plan. Each configuration uses the same 30 recorded decisions and at most a single refresh and replan. Native session tracking detects all 30 same-session changes, while the later executor's observations alone do not expose the inherited plan's old input. Supplying that input through the supported declared-set interface detects all 30 inherited conflicts and restores valid completion. \planfence{} detects every conflict in its completed trials, with a single initialization failure in each stale scenario, and the two separately linked recovery attempts also complete with unchanged settings while the primary table retains the failures. Two initial recorded decisions have task-semantic errors, which accounts for the identical 28 of 30 clean controls.

The distinction is where the evidence persists and where validation takes place. S-Bus validates supplied versions against a single logical bus registry and also supports replicated deployment through Raft, so it is not restricted to a single physical node. In the declared-set configuration, application code supplies the inherited versions through its native interface. \planfence{} instead reconstructs the action's dependency frontier from persistent plan ancestry and queries independently owned public records, which supports plan handoffs across owner-managed memories without a shared bus registry. Both approaches depend on correctly recorded action inputs, and these results compare their evidence requirements and not their transport costs. Recorded-decision completion measures the replayed action's agreement with public requirements and not live-model adaptation, and the observed zero counts are counts over shared templates and not a guarantee over other workloads.

\begin{table}[!htbp]
\centering\small
\setlength{\tabcolsep}{4pt}
{
\begin{tabular}{@{}lrrrr@{}}
\toprule
Method / input evidence & Detected & \shortstack{Invalid\\/ issued} & \shortstack{Valid completion\\/ scheduled} & Infra. \\
\midrule
\multicolumn{5}{l}{\textit{Unchanged input}} \\
S-Bus, session observations & -- & 0/30 & 28/30 & 0 \\
S-Bus, inherited version set & -- & 0/30 & 28/30 & 0 \\
\rowcolor{planfencefill} \planfence{} & -- & 0/30 & 28/30 & 0 \\
\midrule
\multicolumn{5}{l}{\textit{Input revised after same-session planning}} \\
S-Bus, session observations & 30/30 & 0/30 & 30/30 & 0 \\
S-Bus, inherited version set & 30/30 & 0/30 & 30/30 & 0 \\
\rowcolor{planfencefill} \planfence{} & 29/30 & 0/29 & 29/30 & 1 \\
\midrule
\multicolumn{5}{l}{\textit{Inherited plan; executor reads revised input}} \\
S-Bus, session observations & 0/30 & 30/30 & 0/30 & 0 \\
S-Bus, inherited version set & 30/30 & 0/30 & 30/30 & 0 \\
\rowcolor{planfencefill} \planfence{} & 29/30 & 0/29 & 29/30 & 1 \\
\bottomrule
\end{tabular}}
\caption{{Complete native-validator controls. Each configuration schedules 30 recorded-decision trials with at most one replan. Valid completion requires both current derivation inputs and an action satisfying the public requirement. Two initial recorded decisions have task-semantic errors, accounting for the identical 28/30 clean controls. Two \planfence{} initialization failures remain in the stale-case denominators. The comparison concerns evidence and validation, not latency across the native single-service and replicated deployments.}}
\label{tab:validator-comparison}
\end{table}

\FloatBarrier
\section{Connection Reuse}
\label{app:connection-reuse}

Reusing connections reduces transport setup without removing the need to contact an owner. We compare new connections per RPC with persistent connections on the same source and paired workload schedules. The timing slice compares centralized lineage, metadata sync ($K=1$), and \planfence{} at $\rho\in\{0.25,1,4,16\}$ with 8 keys. The scope slice compares batched all-key validation and \planfence{} at 8 and 128 keys with $\rho=4$. The overlapping \planfence{} configurations are run once, yielding 30 configurations across 30 templates. All use five replicas, compact state, a single action dependency, 64 work units, and the AT\&T trace at offset zero. The four nominal update targets yield 28, 21, 11, and 4 protected actions per episode, respectively, and the final update burst is truncated at the horizon where necessary.

Measured sockets open lazily, including their first establishment in the episode, and remain available for subsequent requests. Reuse applies to owner queries, announcements, central-service requests, and content fetches without caching responses or serializing independent owner calls. Each framed exchange still incurs its configured trace delay and service time. The emulator does not add a simulated WAN handshake to TCP connection establishment, so this experiment measures reuse in the implemented transport instead of estimating a carrier's SYN/ACK latency. Traffic counts serialized authenticated frames and excludes TCP/IP headers and retransmissions.

The lower-stall choice still changes with update frequency. With persistent connections centralized lineage still trails \planfence{} by 2.3$\times$ at $\rho=4$ and 4.7$\times$ at $\rho=16$ and leads it at $\rho=0.25$. Figure~\ref{fig:policy-boundary} cell ratios use unrounded medians, and the runner-up in each cell is whichever remaining policy has lower median stall. Bytes and stall can disagree: at $\rho=1$ with reuse \planfence{} sends fewer bytes than metadata sync (8.1 versus 11.6\,KiB per action) yet stalls longer (234 versus 187\,ms), because its owner round trip sits on the action path while announcements do not. Under reuse, the medians of the paired per-template \planfence{} to metadata-sync stall ratios are 2.75, 1.25, 0.56, and 0.19 at $\rho=0.25,1,4,16$, respectively (the ratios of the cohort medians in Table~\ref{tab:connection-reuse} are 2.64, 1.25, 0.56, and 0.19), so the low-churn advantage of metadata sync and the high-churn advantage of action-time validation both persist across the expanded range. The corresponding paired medians against batched all-key validation at $\rho=4$ are 0.84 at 8 keys and 0.62 at 128 keys (ratios of medians 0.85 and 0.63; the 128-key configuration has 29 complete episode pairs). Table~\ref{tab:connection-reuse} gives the episode medians and interquartile ranges, while Figure~\ref{fig:policy-boundary} compares the medians of the lowest-stall and runner-up policies. Reuse leaves framed traffic essentially unchanged while reducing connection establishments for every configuration. At 128 keys, \planfence{} transfers a median 8.1\,KiB per scheduled action compared with 81.7\,KiB for batched all-key validation.

\begin{table}[!htbp]
\centering\footnotesize
\setlength{\tabcolsep}{3pt}
\begin{tabular}{@{}lrrrrrrr@{}}
\toprule
& & & \multicolumn{2}{c}{Stall (ms/action)} & \multicolumn{2}{c}{Traffic (KiB/action)} & Connections \\
\cmidrule(lr){4-5}\cmidrule(lr){6-7}
Method & $|\mathcal{X}|$ & $\rho$ & Fresh & Reused & Fresh & Reused & Fresh / reused \\
\midrule
Centralized lineage & 8 & 0.25 & 151 [147, 166] & 136 [130, 141] & 7.9 & 7.8 & 327 / 20 \\
\shortstack[l]{Metadata sync\\($K=1$)} & 8 & 0.25 & 66 [64, 74] & 55 [51, 57] & 5.8 & 5.8 & 333 / 22 \\
\rowcolor{planfencefill} \planfence{} & 8 & 0.25 & 184 [169, 202] & 146 [140, 157] & 5.8 & 5.8 & 362 / 15 \\
\midrule
Centralized lineage & 8 & 1 & 364 [336, 374] & 308 [303, 337] & 12.3 & 12.3 & 418 / 20 \\
\shortstack[l]{Metadata sync\\($K=1$)} & 8 & 1 & 222 [217, 252] & 187 [184, 190] & 11.6 & 11.6 & 434 / 25 \\
\rowcolor{planfencefill} \planfence{} & 8 & 1 & 306 [274, 327] & 234 [230, 239] & 8.1 & 8.1 & 357 / 15 \\
\midrule
Centralized lineage & 8 & 4 & 566 [553, 613] & 509 [487, 546] & 15.6 & 15.6 & 286 / 19 \\
\shortstack[l]{Metadata sync\\($K=1$)} & 8 & 4 & 483 [467, 534] & 397 [390, 405] & 23.5 & 23.5 & 414 / 25 \\
\shortstack[l]{Batched all-key\\validation} & 8 & 4 & 326 [321, 329] & 262 [260, 265] & 16.4 & 16.4 & 372 / 37 \\
\rowcolor{planfencefill} \planfence{} & 8 & 4 & 244 [242, 248] & 222 [219, 224] & 8.1 & 8.1 & 218 / 13 \\
\midrule
Centralized lineage & 8 & 16 & 1240 [1187, 1308] & 1115 [1105, 1133] & 26.7 & 26.7 & 185 / 17 \\
\shortstack[l]{Metadata sync\\($K=1$)} & 8 & 16 & 1400 [1365, 1557] & 1221 [1193, 1274] & 65.8 & 65.8 & 395 / 25 \\
\rowcolor{planfencefill} \planfence{} & 8 & 16 & 252 [242, 272] & 236 [229, 240] & 8.1 & 8.1 & 120 / 11 \\
\midrule
\shortstack[l]{Batched all-key\\validation} & 128 & 4 & 427 [422, 436] & 352 [349, 363] & 81.7 & 81.7 & 372 / 45 \\
\rowcolor{planfencefill} \planfence{} & 128 & 4 & 243 [241, 254] & 221 [218, 226]$^{\ddagger}$ & 8.1 & 8.1 & 218 / 15 \\
\bottomrule
\end{tabular}
\caption{Connection reuse against a new TCP connection per RPC on the AT\&T trace at offset zero, with five replicas, a single action dependency, and 64 work units. Stall entries give episode medians with interquartile ranges in brackets, and traffic and connection counts are medians. $\ddagger$ marks a configuration with an aborted episode whose cost statistics use the remaining complete episodes. The new-connection column is a separate campaign, on a later harness revision, from Tables~\ref{tab:causal-headline} and~\ref{tab:cadence-appendix}; reuse is compared with new connections within this table only. Connections count establishments during the measured episode, including local client RPCs and excluding proxy-backend duplicates.}
\label{tab:connection-reuse}
\end{table}

\FloatBarrier
All 900 episodes are attempted. A single \planfence{} episode at 128 keys with reuse aborts during an owner update, and the remaining 899 episodes provide 13{,}489 scheduled-action records, including 13{,}487 available actions and two timeout-related blocks. No recorded issued action violates \eqref{eq:current-input}. The aborted episode's 11 scheduled slots remain in the total denominator of 13{,}500, without imputing their unrecorded outcomes or latency. The two blocks occur in fresh \planfence{} at 8 keys and reused batched all-key validation at 128 keys, both at $\rho=4$. The added $\rho=0.25$ and $\rho=16$ configurations complete all 5{,}760 scheduled actions. Two metadata-synchronization RPCs time out within a single persistent-connection episode at $\rho=16$, subsequent barriers restore synchronization before the protected actions, and the timeout costs remain in the measurements. Cost statistics include blocked actions within completed episodes, and the marked configuration has 29 complete episode records instead of 30. The follow-up uses 2{,}000 bootstrap draws resampling paired templates, not individual actions. Reported ordering concerns medians; retained long stalls can make intervals for paired mean differences cross zero. Workers run sequential paired configurations in fixed-seed randomized order, with a single worker per allocated CPU node, and allocations need not occupy exclusive hosts.

\end{document}